\documentclass{article}
\DeclareUnicodeCharacter{0301}{\'{e}}
\usepackage[english]{babel}

\usepackage{setspace}
\usepackage[letterpaper,top=2cm,bottom=2cm,left=2.5cm,right=2.5cm,marginparwidth=1.75cm]{geometry}
\makeatletter
\newcommand{\dashrule}[1][black]{%
  \color{#1}\rule[\dimexpr.5ex-.2pt]{4pt}{.4pt}\xleaders\hbox{\rule{4pt}{0pt}\rule[\dimexpr.5ex-.2pt]{4pt}{.4pt}}\hfill\kern0pt%
}
\newcommand{\rulecolor}[1]{%
  \def\CT@arc@{\color{#1}}%
}
\makeatother
\providecommand{\keywords}[1]
{
  \small	
  \textbf{\textit{Keywords---}} #1
}
\usepackage[sorting = none, backend = biber, style=numeric-comp]{biblatex}
\usepackage{bm}
\usepackage{xurl}
\usepackage{float}
\usepackage{lscape}
\usepackage{lipsum}
\usepackage{caption}
\usepackage{authblk}
\usepackage{verbatim}
\usepackage{rotating}
\usepackage{booktabs}
\usepackage{csquotes}
\usepackage{subfigure}
\usepackage{multirow}
\usepackage{tabularx}
\usepackage{longtable}
\usepackage{graphicx, tikz}
\usetikzlibrary{calc}
\usepackage{amsthm, amsfonts, amssymb}
\usepackage{amsmath, algpseudocode, algorithm}
\usepackage[colorlinks=true, allcolors=blue]{hyperref}
\usetikzlibrary{positioning, fit, arrows.meta, shapes}
\algrenewcommand\algorithmicrequire{\textbf{Inputs:}}
\algrenewcommand\algorithmicensure{\textbf{Output:}}
\usepackage[colorlinks=true, allcolors=blue]{hyperref}

\title{Length of Stay prediction for Hospital Management using Domain Adaptation}
\begin{document}
\def\correspondingauthor{\footnote{Corresponding author: lysenaomi.wambamomo@kuleuven.be\\ \hspace{.2cm} Address: ESAT-STADIUS, Stadius Centre for Dynamical Systems, Signal Processing and Data Analytics, Kasteelpark Arenberg 10, 3001 Leuven, Belgium.}}
\author[a]{Lyse Naomi Wamba Momo \correspondingauthor{}}
\author[b]{Nyalleng Moorosi}
\author[c]{Elaine O. Nsoesie}
\author[d]{Frank Rademakers}
\author[a]{Bart De Moor}
\affil[a]{KU Leuven, Department of Electrical Engineering (ESAT), Stadius Center for Dynamical Systems, Signal Processing and Data Analytics, Leuven, Belgium}
\affil[b]{Google Research}
\affil[c]{Department of Global Health, Boston University School of Public Health}
\affil[d]{Division of Cardiovascular Imaging and Dynamics, Department of Cardiovascular Sciences, KU Leuven, Leuven, Belgium}
\onehalfspacing
\maketitle
\newpage
\begin{abstract}
\noindent Inpatient length of stay (LoS) is an important managerial metric which if known in advance can be used to efficiently plan admissions, allocate resources and improve care. Using historical patient data and machine learning techniques, LoS prediction models can be developed. Ethically, these models can not be used for patient discharge in lieu of unit heads but are of utmost necessity for hospital management systems in charge of effective hospital planning. Therefore, the design of the prediction system should be adapted to work in a true hospital setting.
In this study, we predict early hospital LoS at the granular level of admission units by applying domain adaptation to leverage information learned from a potential source domain. Time-varying data from 110,079 and 60,492 patient stays to 8 and 9 intensive care units were respectively extracted from eICU-CRD and MIMIC-IV. These were fed into a Long-Short Term Memory and a Fully connected network to train a source domain model, the weights of which were transferred either partially or fully to initiate training in target domains. Shapley Additive exPlanations (SHAP) algorithms were used to study the effect of weight transfer on model explanability. 
Compared to the benchmark, the proposed weight transfer model showed statistically significant gains in prediction accuracy (between 1\% and 5\%) as well as computation time (up to 2hrs) for some target domains.
The proposed method thus provides an adapted clinical decision support system for hospital management that can ease processes of data access via ethical committee, computation infrastructures and time.
\\
\end{abstract}

\section{Introduction}
\par Monitoring patients' health condition or recovery trajectory as soon as they are admitted to the hospital or to a critical life-saving unit such as the intensive care unit (ICU) is important to determine and anticipate their needs throughout their entire stay. Patients could transit from a stable to an acutely ill state while under treatment, requiring immediate assistance leading to a prolonged stay. By continuously feeding AI-powered data-driven models, including deep learning methods, with continuously charted patient data (labs, vital signs, medications, prescriptions, etc.), researchers have been able to predict patients' risk of mortality \cite{zhang2020combining, rocheteau2020temporal, rocheteau2021predicting, sheikhalishahi2020benchmarking, harutyunyan2019multitask}, risk of extended length of hospital stay (LoS) \cite{karnuta2020value, karhade2021development}, remaining time in ICU  \cite{song2018attend, harutyunyan2019multitask, khadanga2019using, xu2018raim, rocheteau2020temporal}, hours or days ahead of effective discharge or death. Predictions of LoS or remaining time in ICU would then enable hospital management systems to efficiently allocate both human and logistic resources \cite{rocheteau2020temporal, zolbanin2020data, wu2020development}, reassure family members and more importantly improve patient care and satisfaction \cite{alahmar2018application, zhang2020combining}.
LoS, defined as the duration of inpatient stay from admission to discharge in a single care episode \cite{LOS_defn}, is a tool used to assess the efficiency and effectiveness of hospital management systems. LoS predictions for each individual patient within the hospital at time $t$ would enable timely capacity planning and management.
With the large volumes of data being generated by patients during their hospital stay, accurate, robust and generalizable LoS prediction models can be developed. Robustness being accounted for by the heterogeneous nature of the patient population within an entire hospital.

\par However, the heterogeneity in the patient population that is reflected in the data, for e.g., the difference in the length and density of data points between ICU and non-ICU patients as mentioned in \cite{rajkomar2018scalable}, or in varying frequencies of recording of the same parameters across different units \cite{alves2018dynamic}, different age or disease groups, poses a number of challenges. One of which is the risk of affecting the overall model accuracy especially when the underlying distribution in each sub-population is different.
In order to handle this diversity in patient populations, several authors have restricted their work on specific patient sub-populations delineated by age groups \cite{launay2015predicting, casalino2014predictive}, diagnostics \cite{muhlestein2019predicting} or medical units \cite{ding2009predicting, tsai2016length}. 
In the work presented in \cite{alves2018dynamic}, the authors highlighted this difference in patient behaviors by identifying that PCO2 was more frequently measured in cardiac ICU while Troponin T more frequently in coronary ICU. More specifically, regarding patient heterogeneity and clinical outcome prediction, \cite{suresh2018learning} first clustered ICU patients to obtain more homogeneous groups (clusters) onto which a multitask model was learned for mortality prediction. Multitask learning, much related to transfer learning (TL) \cite{farahani2021brief} was used to simultaneously train and share learned parameters for mortality risk predictions for all clusters. 
Similarly, by considering four ICU sub-populations as different domains (either source or target), domain adaptation was applied in \cite{alves2018dynamic} for fine-tuning pre-trained weights from a CNN-LSTM-FCN network for mortality risk predictions.

\par Domain adaptation is a subclass of TL in which a learned model from a source domain is transferred to a target domain for the prediction of the same task, while accommodating to the difference in data distributions across domains \cite{farahani2021brief}. In the present work, domain adaptation is also applied, for LoS prediction. In both \cite{alves2018dynamic} and \cite{suresh2018learning}, the authors acknowledged the specificities in patient populations for which individual models were constructed using a mechanism (domain adaptation and multitask learning respectively) that also allowed to simultaneously learn the existing similarities in these different populations with a net result in prediction accuracy gains as well as computation time. 
In our work, the specificity of each patient population is emphasized by not restricting the input spaces across all populations to be identical. More specifically, by setting a threshold on the frequency of recording of patient parameters, most recorded features in each population were kept for modelling. Domain adaptation was then applied to transfer pre-trained weights from a source to a target domain, where these two domains could have non-identical input spaces. By so doing, the transfer of pre-trained weights 
only happened for coinciding features between the two domains and for non-coinciding features, random weights were used. As a means to understand this simultaneous training of both pre-trained and random weights, an analysis was carried out by employing discriminative learning \cite{jin2016collaborative} where
different learning rates were assigned to the two sets of features (coinciding and non-coinciding). 
\par In the present work, we employ domain adaptation by transferring knowledge learned from one medical unit or population (source domain) unto others (target domains). Benefits of this approach when transposed to a real hospital environment are multifold because it can help overcome technical challenges related to data modelling and accuracy as well as managerial challenges related to data access and ethical approvals from a hospital. On the technical side, the storage, manipulation, curation, pre-processing and modelling of all patients' data of an entire hospital might stand as a big computing resource challenge in terms of memory capacity and computation time. The latter which might even be more problematic for a real time LoS prediction setup. On the managerial side, obtaining ethical approval for accessing a portion or a unit-level hospital data with addendum for a single unit at a time can be easier and more importantly, faster than requesting one for all patients in the entire hospital. Moreover, given that most hospitals (e.g. in Europe) have been incentivized to work in a very decentralized manner \cite{rechel2018experience}, i.e. allowing decision making at the local hospital unit levels, launching projects like LoS predictions at a unit (smaller) level by the unit head can be more effective and faster than from the hospital top management. 

\par The key contributions of this work are as follows;
\begin{itemize} 
    \item We apply domain adaptation from a source to target domains for hospital LoS predictions for ICU patients using first 24h of data and obtain significant gains in both prediction accuracy and computation time.
    \item By using a second dataset with four potential source domains, we investigate and gain insights into the choice of a suitable source domain for weights initialization on the target domains.
    \item We perform further analysis to understand the relevance and gains of domain adaptation including; discriminative learning to understand the effect of simultaneously training pre-trained and random weights on model accuracy and computation time, the effect of weight transfer on feature importance on the target domains, the transfer of the full model and the use of source domain hyperparameters on all target domains.
\end{itemize}
The remainder of this work is organized as follows; Section \ref{sec:materials_methods}  describes the datasets, the model architecture including domain adaptation, expected gradients method and evaluation metrics. Section \ref{sec:results} covers all results including additional analyses carried out followed by a discussion on these results in Section \ref{sec: discussion}. Finally, in Section \ref{sec: conclusion}, the work is concluded, limitations are discussed and the direction for future work is provided.



\section{Materials and Methods}
\label{sec:materials_methods}
This section discusses the datasets used with descriptive statistics of each ICU unit, the prediction task and the proposed modelling strategies for domain adaptation via weight transfer with and without applying discriminative learning.
\subsection{Datasets}
\label{data_extraction}
The multi-centre eICU collaborative research database (eICU-CRD) \cite{pollard2018eicu} and uni-centre MIMIC-IV \cite{https://doi.org/10.13026/77z6-9w59, goldberger2000physiobank} database were used. By modifying the pipeline by \cite{rocheteau2020temporal}, we selected and pre-processed the first 24 hours of data into ICU. In eICU-CRD, data from 91,277 unique patients corresponding to 110,079 stays admitted to 8 ICU units; Medical Surgical (Med-Surg ICU), Coronary care - cardio-thoracic ICU (CCU-CTICU), Medical ICU (MICU), Neurological ICU (NICU), Cardiac ICU (CICU), Surgical ICU (SICU), Cardio-thoracic ICU (CTICU) and Cardio-surgical ICU (CSICU) was extracted. Similarly, data from 44,245 unique patients and 60,492 ICU stays admitted to 9 ICU units; Medical ICU (MICU), Medical Surgical (Med-Surg ICU), Cardiac Vascular ICU (CVICU), Surgical ICU (SICU), Trauma ICU (TICU), Coronary care ICU (CCU), Neuro surgical ICU (Neuro SICU), Neuro intermediate (NI), Neuro stepdown (NS) was extracted from MIMIC-IV. After hourly sampling extracted data using the mean, features were only kept for modelling if at least 2 unique recordings over 24h were present for at least 30\% of the patients.\footnote{see Appendix \ref{sec:extracted_features} for details on feature extraction}.
Missing values were then imputed by first forward filling and then backward filling the most recent value for each patient, as in \cite{purushotham2018benchmarking, che2018recurrent}.
\begin{table}[H]
    \caption[The input space was augmented with binary indicators of the same size as the original input space to indicate imputed and newly recorded parameter values. A variable \textit{hour} indicating the time of the day at which a record was done was also added.]{Baseline Characteristics per ICU unit in eICU-CRD. Total No. features = (No.Inputs $\times$ 2)+1 \footnotemark}
    \centering
    \begin{tabular}{p{2.5cm}p{2cm}p{2cm}p{2cm}p{1.5cm}p{2cm}p{1.5cm}}
    \toprule
    ICU unit &No. ICU stays & No.Inputs & Mean(LoS) & Std(LoS) & Median(Age) & Gender (\% Male)\\
    \midrule
     Med-Surg ICU & 58,335 & 25 & 7.55 & 6.88 & 66 & 52.9\\
     MICU & 10,128 & 24  &9.24 &8.65 & 65 & 52.1\\     
     CCU-CTICU & 9,950 & 27 & 7.24 & 6.78 & 67 & 59.0 \\
     NICU & 8,777 & 25  &8.26 & 8.04 & 63 & 51.9\\
     CICU & 7,744 & 25 & 7.87 & 7.61 & 65 & 50.4 \\
     SICU & 7,684 & 26 &9.45 & 8.47 & 65 & 57.5\\
     CTICU & 4,286 & 33  & 8.79 &7.58 & 66 & 62.5 \\
     CSICU & 3,175 & 35  & 7.32 & 6.39 & 69 & 59.8\\
     \hline
     All stays & 110,079 & 26 & 7.93 & 7.36 &65 & 54.3\\
     \bottomrule
    \end{tabular}
    \label{tab:eicu}
    \end{table}
\footnotetext{The input space was augmented with binary indicators of the same size as the original input space to indicate imputed and newly recorded parameter values following \cite{lipton2016modeling}. A variable \textit{hour} indicating the time of the day at which a record was done was also added.}

\begin{table}[H]
    \caption{Baseline Characteristics per ICU unit in MIMIC-IV. Total No. features = (No.Inputs $\times$ 2)+1}
    \label{tab:mimiciv}
    \centering
    \begin{tabular}{p{2.5cm}p{2cm}p{2cm}p{2cm}p{1.5cm}p{2cm}p{1.5cm}}
    \toprule
    ICU unit &No. ICU stays & No.Inputs & Mean(LoS) & Std(LoS) & Median(Age) & Gender (\% Male)\\
    \midrule
     MICU & 12,378 & 29 & 9.88 & 8.99 & 65 & 54.4 \\
     CVICU & 11,070 & 38 & 8.05 & 7.03 & 69 & 67.4 \\
     Med-Surg ICU & 10,152 & 23 & 9.93 & 9.41 & 66 & 51.3 \\
     SICU & 9,262 & 28 & 10.82 & 9.94 & 64 & 53.5 \\
     TSICU & 6,992 & 35 & 7.87 & 7.61 & 64 & 57.6 \\
     CCU & 6,881 & 24 & 8.36 & 7.72 & 72 & 57.1\\
     Neuro SICU & 1,646 & 28 & 11.71 & 10.75 & 67 & 50.6\\
     NI & 1,491 & 20 & 7.15 & 7.67 & 68 & 50.6 \\
     NS & 620 & 20 & 7.53 & 7.46 & 68 & 52.6 \\
     \hline
     All stays & 60,492 & 33 & 9.69 & 9.02 & 67 & 56.5\\
     \bottomrule
    \end{tabular}
\end{table}
\subsection{Prediction task}
In this study, we predict the time lapse between ICU admission and hospital discharge using the first 24h of data into ICU. Selected admissions in both datasets are such that they have spent at least 24h in ICU to prevent data leakage. Datasets were split into train, validation and test sets following the 70:15:15 ratio. 
\subsection{Model architecture}
The model architecture used consists of an LSTM layer followed by a fully connected layer (FCN) to handle the temporal dimension in the data and output the LoS prediction (in fractional days) respectively. 
\subsubsection{Long-Short Term Memory neural network (LSTM)}
\begin{figure}[H]
\centering
\begin{tikzpicture}[
    font=\sf \scriptsize,
    >=LaTeX,
    cell/.style={
        rectangle, 
        rounded corners=5mm, 
        draw,
        very thick,
        },
    operator/.style={
        circle,
        draw,
        inner sep=-0.5pt,
        minimum height =.2cm,
        },
    function/.style={
        ellipse,
        draw,
        inner sep=1pt
        },
    ct/.style={
        circle,
        draw,
        line width = .75pt,
        minimum width=1cm,
        inner sep=1pt,
        },
    gt/.style={
        rectangle,
        draw,
        minimum width=4mm,
        minimum height=3mm,
        inner sep=1pt
        },
    mylabel/.style={
        font=\scriptsize\sffamily
        },
    ArrowC1/.style={
        rounded corners=.25cm,
        thick,
        },
    ArrowC2/.style={
        rounded corners=.5cm,
        thick,
        },
    block/.style={draw=gray,dashed,fill=green!1,thick,inner sep=5pt},
    ]
\node [cell, minimum height=4cm, minimum width=8cm] at (0,0){};

\node [gt] (ibox1) at (-2.5,-0.75) {$\sigma$};
 \node [gt] (ibox2) at (-1.5,-0.75) {$\sigma$};
 \node [gt, minimum width=1cm] (ibox3) at (-0.5,-0.75) {ReLU};
 \node [gt] (ibox4) at (0.7,-0.75) {$\sigma$};
 
\draw[red,thick,dotted] ($(ibox1.north west)+(-0.65,3.0)$)  rectangle ($(ibox1.south east)+(0.2,-1)$);
\draw[red,thick,dotted] ($(ibox2.north west)+(-0.15,3.0)$)  rectangle ($(ibox3.south east)+(0.15,-1)$);

\node [label={[align=left] previous states}] (ft) at (-5.2, 2.3) {};
\node [label={[align=left] forget gate}] (ft) at (-2.8, 2.3) {};
\node [label={[align=left] input gate}] (ft) at (-0.8, 2.3) {};
\node [label={[align=left] output gate}] (ft) at (1.2, 2.3) {};

\node [operator] (mux1) at (-2.5,1.5) {$\times$};
\node [operator] (add1) at (-0.5,1.5) {+};
\node [operator] (mux2) at (-0.5,0) {$\times$};
\node [operator] (mux3) at (1.5,0) {$\times$};
\node [function] (func1) at (1.5,0.75) {ReLU};

\draw[red,thick,dotted] ($(ibox4.north west)+(-0.15,3.0)$)  rectangle ($(func1.south east)+(0.27,-2.53)$);

\node [ct, label={[mylabel]cell state}] (c) at (-5,1.5) {$c^{\langle t-1 \rangle}$};
\node[ct, label={[mylabel]hidden}] (h) at (-5,-1.5) {$h^{\langle t-1 \rangle}$};
\node[ct, label={[mylabel]below:Input}] (x) at (-2.85,-3) {$x^{\langle t \rangle}$};

\draw[red,thick,dotted] ($(c.north west)+(-0.75,0.55)$)  rectangle ($(c.south east)+(0.5,-3.2)$);

\node[ct, label={[mylabel]}] (c2) at (5,1.5) {$c^{\langle t \rangle}$};
\node[ct, label={[mylabel]}] (h2) at (5,-1.5) {$h^{\langle t \rangle}$};
\node[ct, label={[mylabel]left:}] (x2) at (2.5,3) {$h^{\langle t \rangle}$};
  
\draw [->, ArrowC1] (c) -- (mux1) -- (add1) -- (c2);

\draw [ArrowC2] (h) -| (ibox4);
\draw [ArrowC1] (h -| ibox1)++(-0.5,0) -| (ibox1); 
\draw [ArrowC1] (h -| ibox2)++(-0.5,0) -| (ibox2);
\draw [ArrowC1] (h -| ibox3)++(-0.5,0) -| (ibox3);
\draw [ArrowC1] (x) -- (x |- h)-| (ibox3);

\draw [->, ArrowC2] (ibox1) -- (mux1) node[midway, right] {$f_t$};
\draw [->, ArrowC2] (ibox2) |- (mux2) node[midway]{$i_t$};
\draw [->, ArrowC2] (ibox3) -- (mux2) node[midway,left]{$\tilde{c}_t$};
\draw [->, ArrowC2] (ibox4) |- (mux3) node[midway]{$o_t$};
\draw [->, ArrowC2] (mux2) -- (add1);
\draw [->, ArrowC1] (add1 -| func1)++(-0.5,0) -| (func1);
\draw [->, ArrowC2] (func1) -- (mux3);

\draw [->, ArrowC2] (mux3) |- (h2);
\draw (c2 -| x2) ++(0,-0.1) coordinate (i1);
\draw [-, ArrowC2] (h2 -| x2)++(-0.5,0) -| (i1);
\draw [->, ArrowC2] (i1)++(0,0.2) -- (x2);
\end{tikzpicture}
\caption{LSTM architecture with forget gate \cite{gers2000learning}}
\label{LSTM_architec}
\end{figure}
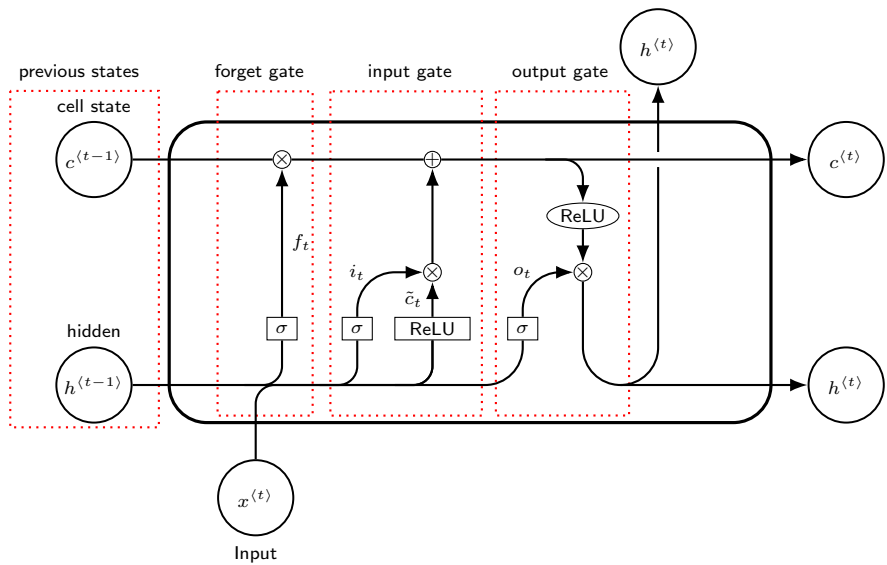
LSTM networks \cite{hochreiter1997long} belong to the family of RNNs \cite{robinson1987utility, werbos1988generalization, ranzato2014video, yu2019review} used to model and learn long-term dependencies from sequential data.
The structure used in this work is given in Figure \ref{LSTM_architec}.
At the core of RNNs like LSTM is the recurrent cell with its memory cell state $c_t$ that holds previous states and current input information. The recurrent cell of the LSTM in Figure \ref{LSTM_architec} is given by; \begin{equation}
\begin{aligned}
\tilde{c}_t &= g\left(W_{\tilde{c}h}h_{t-1} + W _{\tilde{c}x}x_t + b_{\tilde{c}}\right) \\
c_t &= f_t \cdot c_{t-1} + i_t \cdot \tilde{c}_t \\
h_t &= o_t \cdot g(c_t),
\label{cellstate}
\end{aligned}
\end{equation}
where $h_t$ is the hidden state at a discrete time $t$ (here, $t=1$ hour), $W_{ab}$ are weight parameters from layers $a$ to $b$, $\cdot$ is the element-wise product and $g()$, a non-linear activation function applied to the results of matrix operations. Here, $g()$ was taken as the rectified linear Unit (ReLU) \cite{jozefowicz2015empirical};
\begin{equation}
    ReLU(x) = max(0, x).
\end{equation}
The gate functions $f_t, o_t$ and $i_t$ control the amount of information flow via a sigmoid (logistic) activation function. These gates are respectively the forget, the output and the input and are defined as;
\begin{equation}
\begin{aligned}
 i_t &= \sigma\left(W_{ih}h_{t-1}+W_{ix}x_t+b_i\right)\\
 f_t &= \sigma\left(W_{fh}h_{t-1}+W_{fx}x_t+b_f\right)\\
 o_t &= \sigma\left(W_{oh}h_{t-1}+W_{ox}x_t+b_o\right).
 \label{gates}
\end{aligned}
\end{equation}
For implementation, we used the open source deep learning (DL) library, Keras \cite{chollet2015keras} with Tensorflow \cite{abadi2016tensorflow} as back-end.
In Keras, the weights $W_{ab}$ are distributed in the kernel, the recurrent kernel and the bias term of the LSTM layer. The kernel stores all weights multiplied by the inputs $x_t$, i.e., $\bm{W_x} = \{W_{\tilde{c}x}, W_{ix}, W_{fx}, W_{ox}\}$, the recurrent kernel, those multiplied by the hidden state $h_{t-1}$, i.e., $\bm{W_h} = \{W_{\tilde{c}h}, W_{ih}, W_{fh}, W_{oh}\}$ and the bias stores all bias terms, i.e., $\bm{b} = \{b_{\tilde{c}}, b_i, b_f, b_o\}$.
Therefore, the kernel weight matrix is of shape $(n_D\times 4H)$, where $n_D$ is the input dimension of the domain $D$ (source or target) and $H$, the number of hidden units of the LSTM layer. The recurrent weight matrix is of shape $(H \times 4H)$ and the bias is a vector of shape $(4H \times 1)$.\\
From equations (\ref{cellstate}) and (\ref{gates}), it follows that the model generates a prediction at each time-step $t$ given by $h_t$. This structure sometimes referred to as a many-to-many LSTM architecture was not used in this work, rather the many-to-one structure, in order to get only the prediction at the last time-step ($t=24h$).
\subsubsection{Fully connected network (FCN)}
Given the last hidden state $h_t$ from the LSTM layer, the final LoS prediction was computed as; 
\begin{equation}
    \hat{y}_t = g\left(W_{y_t}h_t + b_{y_t}\right),
\end{equation}
with $g$ again taken as the ReLU to output positive predictions. This fully connected layer contains only one unit to perform LoS regression and output a unique value for LoS.
\subsection{Domain adaptation}
\subsubsection{Domain adaptation via weight transfer}
The literature has shown that the use of pre-trained weights can be more beneficial than using random initial weights, sometimes, regardless of the difference in prediction tasks \cite{yosinski2014transferable} and even from one related dataset to another \cite{pmlrv56Choi16, lee2017transfer}. The net benefit of fine-tuning these pre-trained weights on the target domain is often computation time for hyperparameters optimization and model training \cite{gupta2020transfer}.\\
Given the input space $\bar{\bm{X}}$ extracted from the database, the model inputs $\bm{X}$ are obtained by augmenting this input matrix with a matrix of binary indicators $\tilde{\bm{X}}$ such that, $\bm{X} = \text{concat}(\bar{\bm{X}}; \tilde{\bm{X}})$ where,
\begin{equation}
\tilde{\bm{X}} = \begin{cases}
0, if \hspace{.1cm}\bar{\bm{X}} \text{is recorded} \\
1, if \hspace{.1cm} \bar{\bm{X}} \text{is imputed},
\end{cases}
\end{equation}
following \cite{lipton2016modeling}.\\
Following feature extraction in section \ref{data_extraction}, the input matrix $\bm{X}$ can differ both in the number of features and/or the features themselves (see, Tables \ref{tab:eicu} and \ref{tab:mimiciv} and Appendix \ref{sec:extracted_features}) between units. Thus given a source ($S$) and a target ($T$), we have; 
\begin{subequations}
\begin{align}
    \bm{X}_S &\subset \bm{X}_T \label{eqn:subset1}\\
    \bm{X}_S &\not\subset \bm{X}_T \hspace{.2cm} \text{or} \hspace{.2cm} \bm{X}_T \not\subset \bm{X}_S \label{eqn:subset2}\\
    \bm{X}_S &= \bm{X}_T \hspace{.2cm} \text{or} \hspace{.2cm} \bm{X}_T \subset \bm{X}_S \label{eqn:equal}
    \end{align}
    \label{input_spaces}
\end{subequations}
In the case of (\ref{eqn:subset1}) and (\ref{eqn:subset2}), only partial weight transfer from the source to the target can occur for coinciding features between the two domains.  In case (\ref{eqn:equal}), where all inputs in $T$ are found in $S$, total weight transfer occurs. The proposed architecture is given in Figure \ref{fig:weight_transfer}.
\tikzset{%
  every neuron/.style={
    circle,
    draw,
    minimum size=0.5cm
  },
  neuron missing/.style={
    draw=none, 
    scale=1.2,
    text height=0.15cm,
    execute at begin node=\color{black}$\vdots$
  },
  weightscoinc/.style = {
  circle, 
  draw, 
  minimum size=1cm,
  path picture={\fill
  (0:0em) circle[radius=2.0pt] (-1:-1em) circle[radius=2.0pt] (-260:0.5em) circle[radius=2.0pt] 
  (140:0.9em) circle[radius=2.0pt] (2.2:-400em) circle[radius=2.0pt]
;}
  },
 weightsnoncoinc/.style = {
 circle, 
 draw, 
  minimum size=1cm,
  path picture={\fill 
  (140:1.3em) circle[radius=2.0pt] (100:0.9em) circle[radius=2.0pt] (150:0.9em) circle[radius=2.0pt] (200:0.9em) circle[radius=2.0pt]
  ;}
 },
}
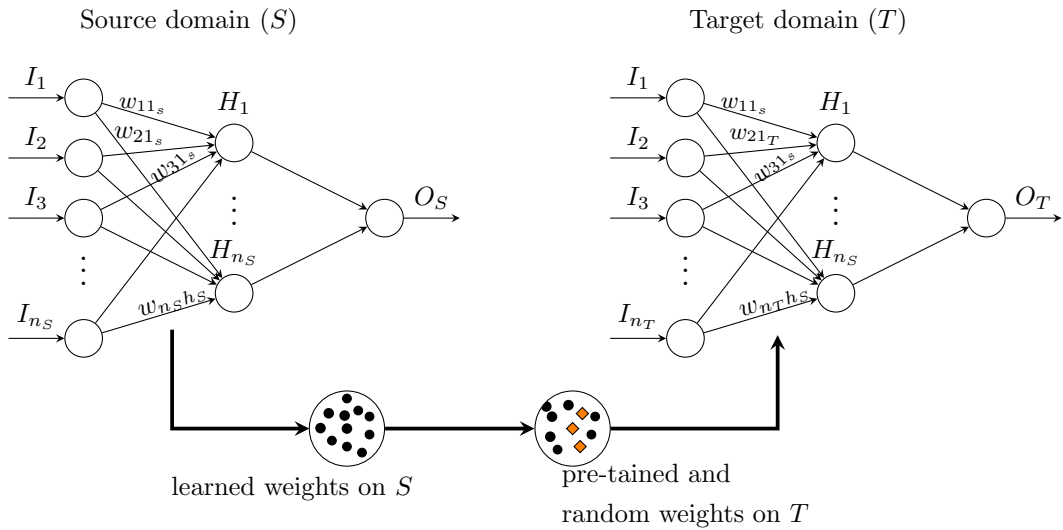
\begin{figure}[H]
    \centering
\begin{tikzpicture}[x=1.0cm, y=0.8cm, >=stealth]

\foreach \m/\l [count=\y] in {1,2,3,missing,4}
  \node [every neuron/.try, neuron \m/.try] (input-\m) at (0,2.5-\y) {};

\foreach \m [count=\y] in {1,missing,2}
  \node [every neuron/.try, neuron \m/.try ] (hidden-\m) at (2,2-\y*1.25) {};

\node [every neuron] (output) at (4,-0.5) {}; 

\foreach \l [count=\i] in {1,2,3,n_S}
  \draw [<-] (input-\i) -- ++(-1,0)
    node [above, midway] {$I_{\l}$};

\foreach \l [count=\i] in {1,n_S}
  \node [above] at (hidden-\i.north) {$H_{\l}$};

\draw [->] (output) -- ++(1,0) node [above, midway] {$O_S$};

\foreach \i in {1,...,4}
  \foreach \j in {1,...,2}
    \draw [->] (input-\i) -- (hidden-\j);

\node [rotate=30, label={[align=center] \small $w_{{11}_s}$}] at (1.3, 0.95) {};
\node [rotate=45, label={[align=right] $w_{{21}_s}$}] at (1.3, 0.45) {};
\node [rotate=45, label=above:] at (1.2, 0.38) {$w_{{31}_s}$};
\node [rotate=20, label=above:] at (1.2, -1.9) {$w_{n_Sh_S}$};

\foreach \i in {1,...,2}
    \draw [->] (hidden-\i) -- (output);

 \node [align=center, above] at (1.4, 2.4) {Source domain ($S$)};
 \node(coinc) [anchor=east] at (1.3,-2.2) {};
 \node(ncoinc)[anchor=east] at (3.0,-4){};
 \draw[->, line width=0.5mm] (coinc.south) |- (ncoinc.east);
 \node (wc) [weightscoinc] at (3.5,-4) {};
 \node(pretrainw) [anchor=east] at (4.5,-5.0){learned weights on $S$};
 \node(wnc) [weightsnoncoinc] at (6.5,-4) {} ;
 \draw[->, line width=0.5mm] (wc) -- (wnc);
 \node(ncoincnet) [anchor=east] at (9.35,-2.6) {};
 \draw[->, line width=0.5mm] (wnc.east) -| (ncoincnet.north);
 \node(randomw) [label={[align=left] pre-tained and \\random weights on $T$}] at (8.0, -5.95){};
 \node[draw, circle, minimum size=0.05cm, scale=0.35, fill=black] at (3.8,-3.8){};
  \node[draw, circle, minimum size=0.05cm, scale=0.35, fill=black] at (3.8,-4.1){};
  \node[draw, circle, minimum size=0.05cm, scale=0.35, fill=black] at (3.5,-4.3){};
  \node[draw, circle, minimum size=0.05cm, scale=0.35, fill=black] at (3.3,-4.2){};
  \node[draw, circle, minimum size=0.05cm, scale=0.35, fill=black] at (3.5,-3.5){};
\node[draw, circle, minimum size=0.05cm, scale=0.35, fill=black] at (3.65,-3.7){};
\node[draw, circle, minimum size=0.05cm, scale=0.35, fill=black] at (3.7,-4.4){};
 \node[draw, diamond, minimum size=0.05cm, scale=0.35, fill=orange] at (6.5,-4){};
  \node[draw, diamond, minimum size=0.05cm, scale=0.35, fill=orange] at (6.6,-4.3){};
\node[draw, circle, minimum size=0.05cm, scale=0.35, fill=black] at (6.8,-3.8){};
 \node[draw, circle, minimum size=0.05cm, scale=0.35, fill=black] at (6.75,-4.1){};
  \node[draw, circle, minimum size=0.05cm, scale=0.35, fill=black] at (6.3,-4.35){};
\node[draw, diamond, minimum size=0.05cm, scale=0.35, fill=orange] at (6.63,-3.75){};

\foreach \m/\l [count=\y] in {1,2,3,missing,4}
  \node [every neuron/.try, neuron \m/.try] (input-\m) at (8.0,2.5-\y) {};
\foreach \m [count=\y] in {1,missing,2}
  \node [every neuron/.try, neuron \m/.try ] (hidden-\m) at (10,2-\y*1.25) {};
\node [every neuron] (output) at (12,-0.5) {};

\foreach \l [count=\i] in {1,2,3,n_T}
  \draw [<-] (input-\i) -- ++(-1,0)
    node [above, midway] {$I_{\l}$};

\foreach \l [count=\i] in {1,n_S}
  \node [above] at (hidden-\i.north) {$H_{\l}$};

\draw [->] (output) -- ++(1,0) node [above, midway] {$O_T$};

\foreach \i in {1,...,4}
  \foreach \j in {1,...,2}
    \draw [->] (input-\i) -- (hidden-\j);

\node [rotate=30, label={[align=center] \small $w_{{11}_s}$}] at (9.3, 0.95) {};
\node [rotate=45, label={[align=right] \small $w_{{21}_T}$}] at (9.5, 0.45) {};
\node [rotate=45, label=above:] at (9.2, 0.38) {\small $w_{{31}_s}$};
\node [rotate=20, label=above:] at (9.2, -1.9) {$w_{n_Th_S}$};

\foreach \i in {1,...,2}
    \draw [->] (hidden-\i) -- (output);
    
\node [align=center, above] at (9.5, 2.4) {Target domain ($T$)};
\end{tikzpicture}
    \caption{Proposed architecture: Domain adaptation via weight transfer from the source domain to the target domains. $I_{n_S}$, $I_{n_T}, H_{n_S}, O_S, O_T$ are respectively, the source inputs, target inputs, source hidden unit, source output and target output and ${n_S}\neq {n_T}$. Black dots are weights of coinciding features and orange diamonds are random weights for non-coinciding features.}
    \label{fig:weight_transfer}
\end{figure}
\noindent The pseudo algorithm used for assigning pre-trained weights (either partially or totally) from the source to the target is given in Algorithm \ref{alg:weights_transfer}.
\begin{algorithm}
\caption{Weight Transfer from Source domain $S$ to Target domain $T$}\label{alg:weights_transfer}
\begin{algorithmic}
\Require Given:\\
Target inputs of size $m_T\times 24 \times n_T$: $\bm{X}_T = \{x_{1t}, ..., x_{n_{T_t}}\}$, \hspace{.05cm} $t=1,...,24$ \\
Source inputs of size $m_S\times 24 \times n_S$: $\bm{X}_S = \{x_{1t}, ..., x_{n_{S_t}}\}$, \hspace{.05cm} $t=1,...,24$ \\
$W_{xS}$= load source model kernel weights of size $N_S \times 4H$
\Ensure Target kernel weight matrix $W_{xT}$ of size $N_T \times 4H$
\State Initialization of weight matrix $W_{xT}$ using Glorot uniform distribution \cite{glorot2010understanding}
\For{$x \in \bm{X}_T$}
\State target\_index $=\bm{X}_T[x]$
\If{$x \in \bm{X}_S$}
\State $W_{xT}$[target\_index] = $W_{xS}$[source\_index]
\Else
\State $W_{xT}[x]$ = $W_{xT}[x]$
\EndIf
\EndFor
\end{algorithmic}
\end{algorithm}
\noindent After initializing the LSTM model for each $T$, model weights are set using the fully trained model from S and/or random kernel weights following Algorithm \ref{alg:weights_transfer}, recurrent kernel and bias weights from $S$, i.e., $[\bm{W}_{xT}, \bm{W}_{hS}, \bm{b}_S]$.
\subsubsection{Domain Adaptation via weight transfer and discriminative learning}
\label{sec:diff_lr}
\begin{figure}[H]
\centering
\includegraphics[width=.5\linewidth]{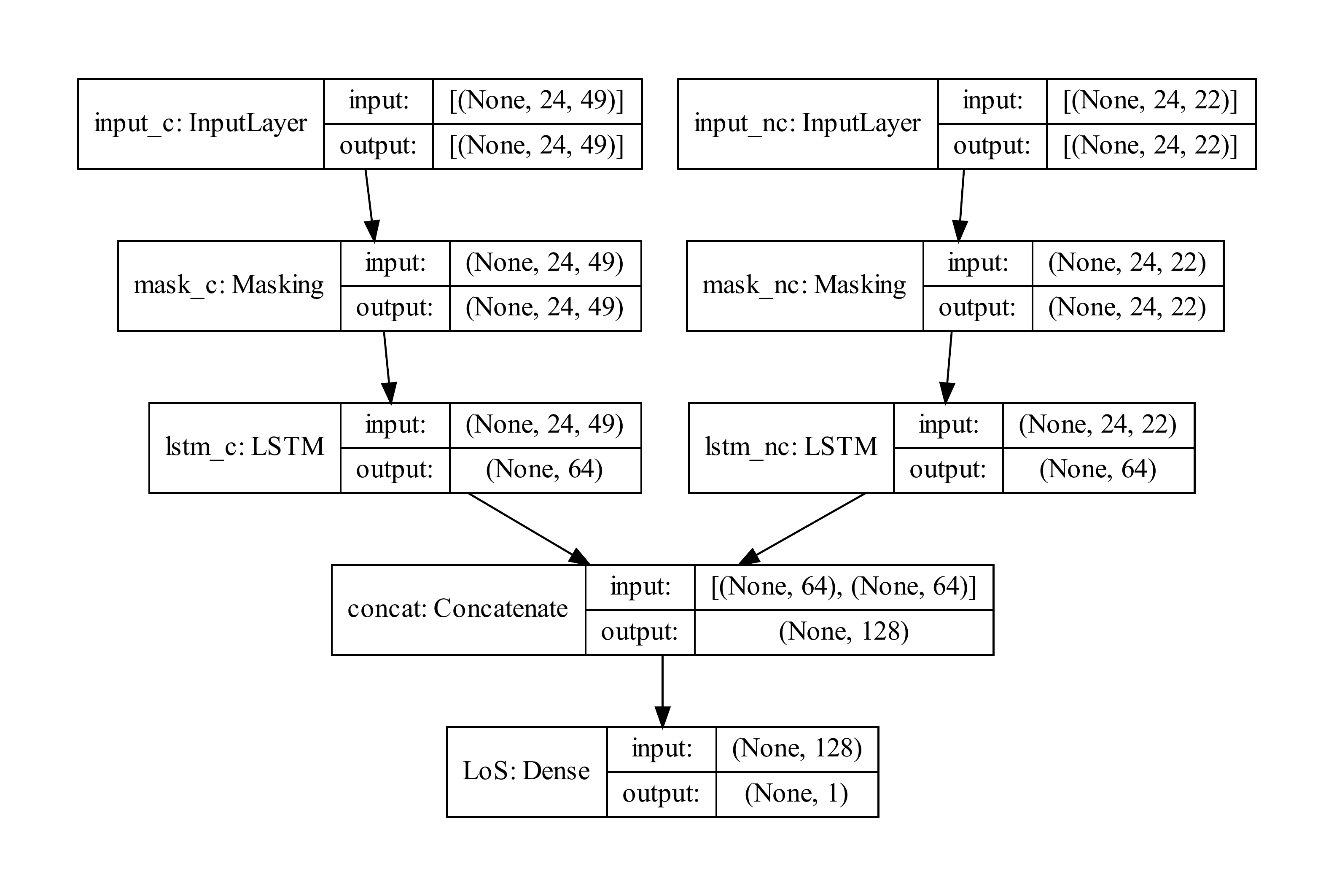}
\caption{Transfer learning model with different learning rates assigned to layers \textit{lstm\_c} (layer with coinciding features) and \textit{lstm\_nc} (layer with non-coinciding features) for CSICU in eICU dataset}
\label{fig:diff_lr_csicu}
\end{figure}

During TL, all hyperparameters from $S$ except the batch size are transferred, i.e., the learning rate, the dropout rate and the number of hidden units from $S$ are all transferred to $T$. Our intuition here is that during training, coinciding features between $T$ and $S$ that receive fully trained weights are overfitting whilst the random weights of non-coinciding features are still learning. To study this, discriminative fine-tuning was applied, (see Figure \ref{fig:diff_lr_csicu}), s.t., different learning rates ($lr$) are assigned to the two groups of inputs (layers) and optimized using a multi-optimizer \cite{howard2018universal, jin2016collaborative}.
\begin{equation}
    lr_T = 
    \begin{cases}
lr_S \hspace{.3cm} if \hspace{.3cm} {X}_T  \not\in  \bm{X}_S \\
\alpha lr_S \hspace{.3cm} if \hspace{.3cm}  \bm{X}_S \equiv \bm{X}_T
    \end{cases}
    \label{eqn:diff_lr}
\end{equation}
The learning rate from $S$ is transferred to the non-coinciding features (features in $T$ not in $S$) while it is reduced by a power of $\alpha$ for coinciding features. 
\subsection{Interpretability using Expected gradients:}
\label{XAI_EG_SHAP}
Expected gradients (EG) method \cite{erion2021improving} were used in order to unlock the black-box nature of our LSTM models and appreciate the effect of weights transfer on overall feature importance. The gradient explainer of the SHAP software package \footnote{\url{https://github.com/slundberg/shap}} was used. Ample details on the method can be found in \cite{erion2021improving}.
\subsection{Evaluation metrics:}
Model training was carried out using the mean squared log error (MSLE) to take into account the skewness in the outcome and model performance reported using the following metrics, \[MAE = \frac{1}{n_D}\sum _{i=1}^{n_D}|y_i-\hat{y_i}|, MAPE=\frac{1}{n_D}\sum _{i=1}^{n_D}\bigg|\frac{y_i-\hat{y_i}}{y_i}\bigg|, MSE=\frac{1}{n_D}\sum _{i=1}^{n_D}(y_i-\hat{y_i})^{2},\] where $n_D$ is the number of stays in each domain. 
\section{Results}
This section discusses all numerical results without and upon applying weights transfer, including some further analyses such as no hyperparameter optimization on target domains, full model transfer and discriminative learning. 
\subsection{Hyperparameter Optimization}
\label{hyperpar_optim}
\begin{figure}[H]
    \centering
    \subfigure[Med-Surg ICU]{
\label{fig:med_surg_loss}
        \includegraphics[width=.27\linewidth]{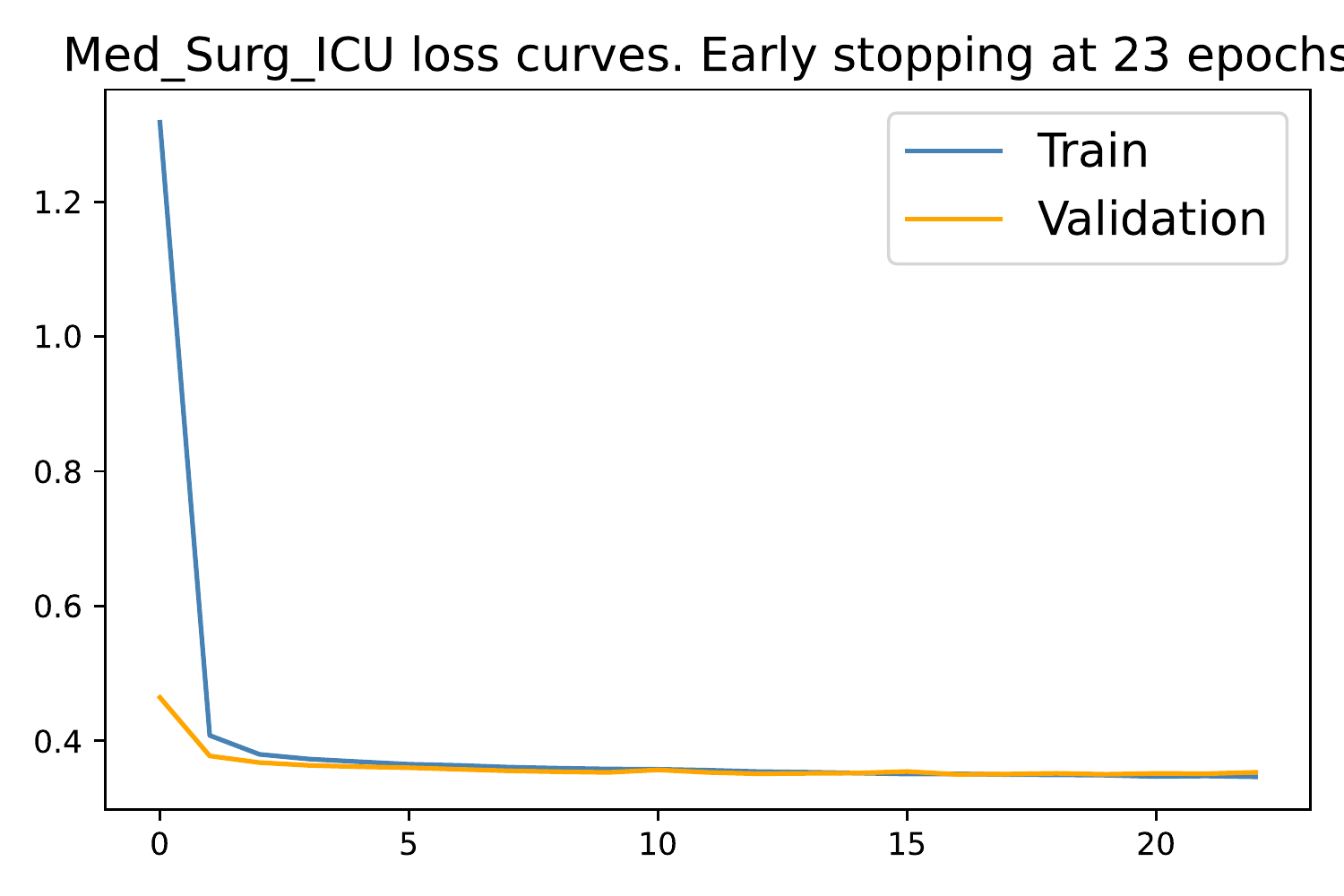}}
        \hspace{1cm}
    \subfigure[CTICU]{
\label{fig:cticu_loss}
        \includegraphics[width=.27\linewidth]{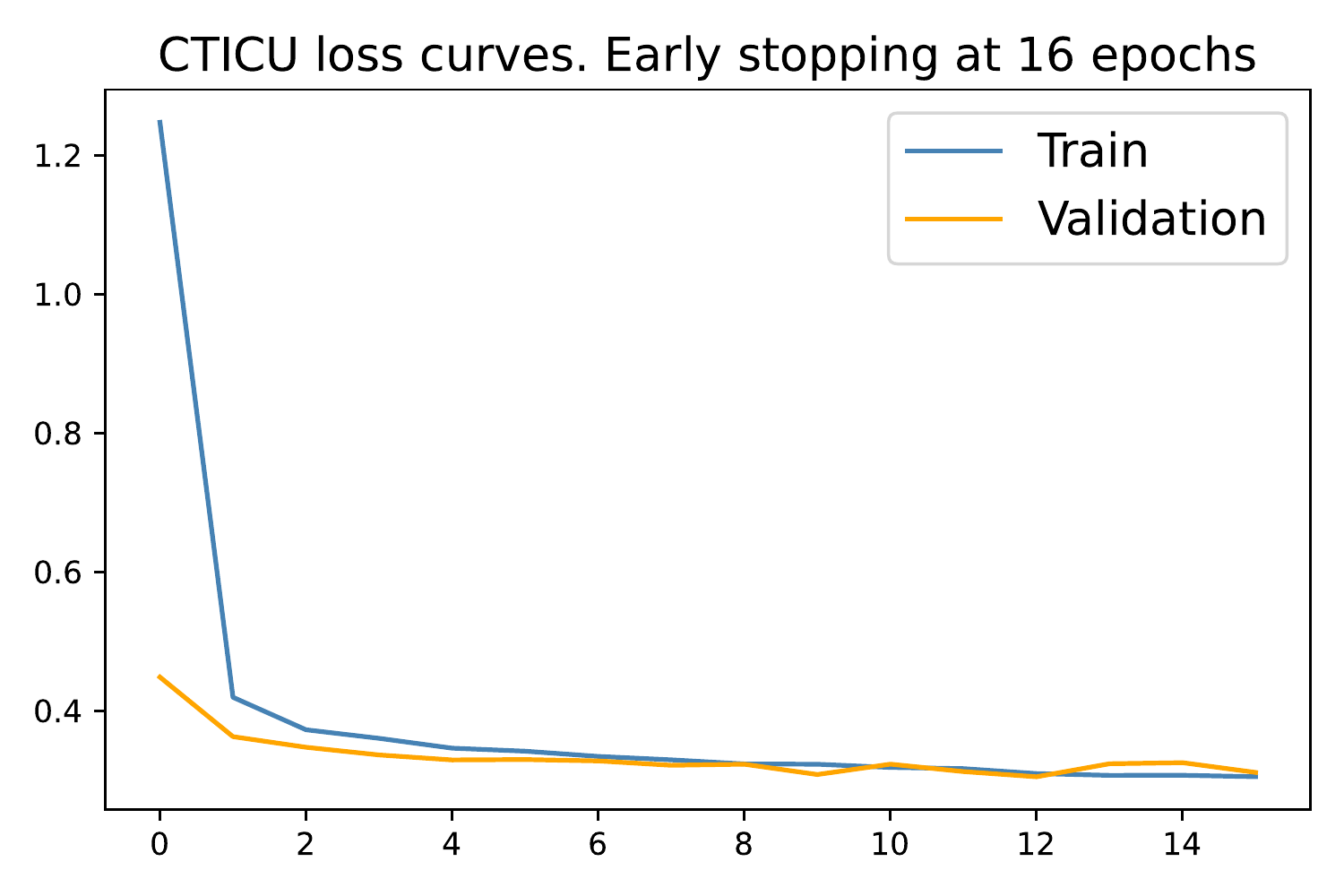}}
        \hspace{1cm}
    \subfigure[MICU]{
\label{fig:micu_loss}
        \includegraphics[width=.27\linewidth]{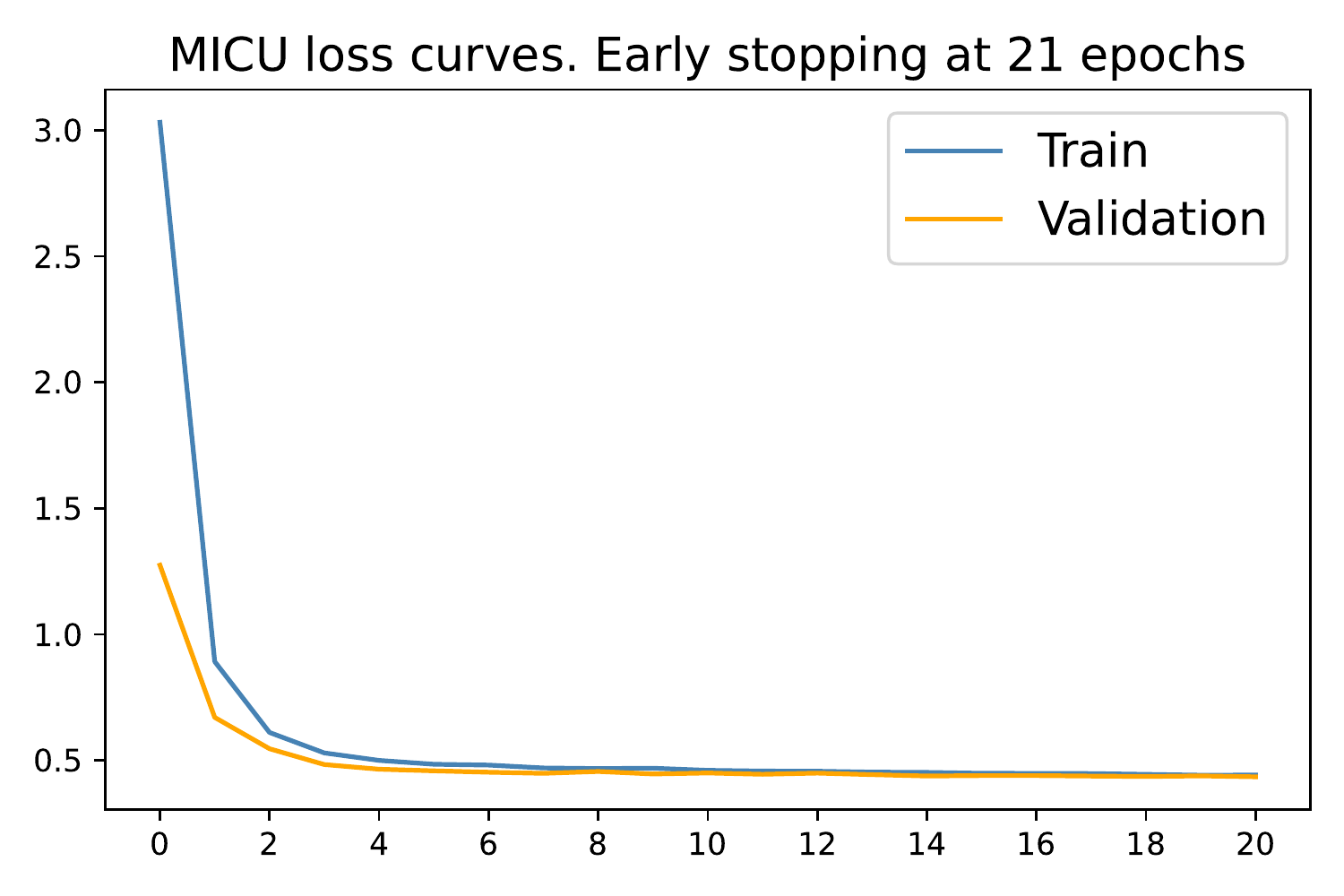}}
        \caption{Training and Validation curves obtained after model optimization on some eICU-CRD units.}
    \label{fig:eicu_loss_curves}
\end{figure}

\begin{figure}[h]
    \centering
    \subfigure[All patients data]{
 \label{fig:all_pat_loss_mimic}
        \includegraphics[width=.27\linewidth]{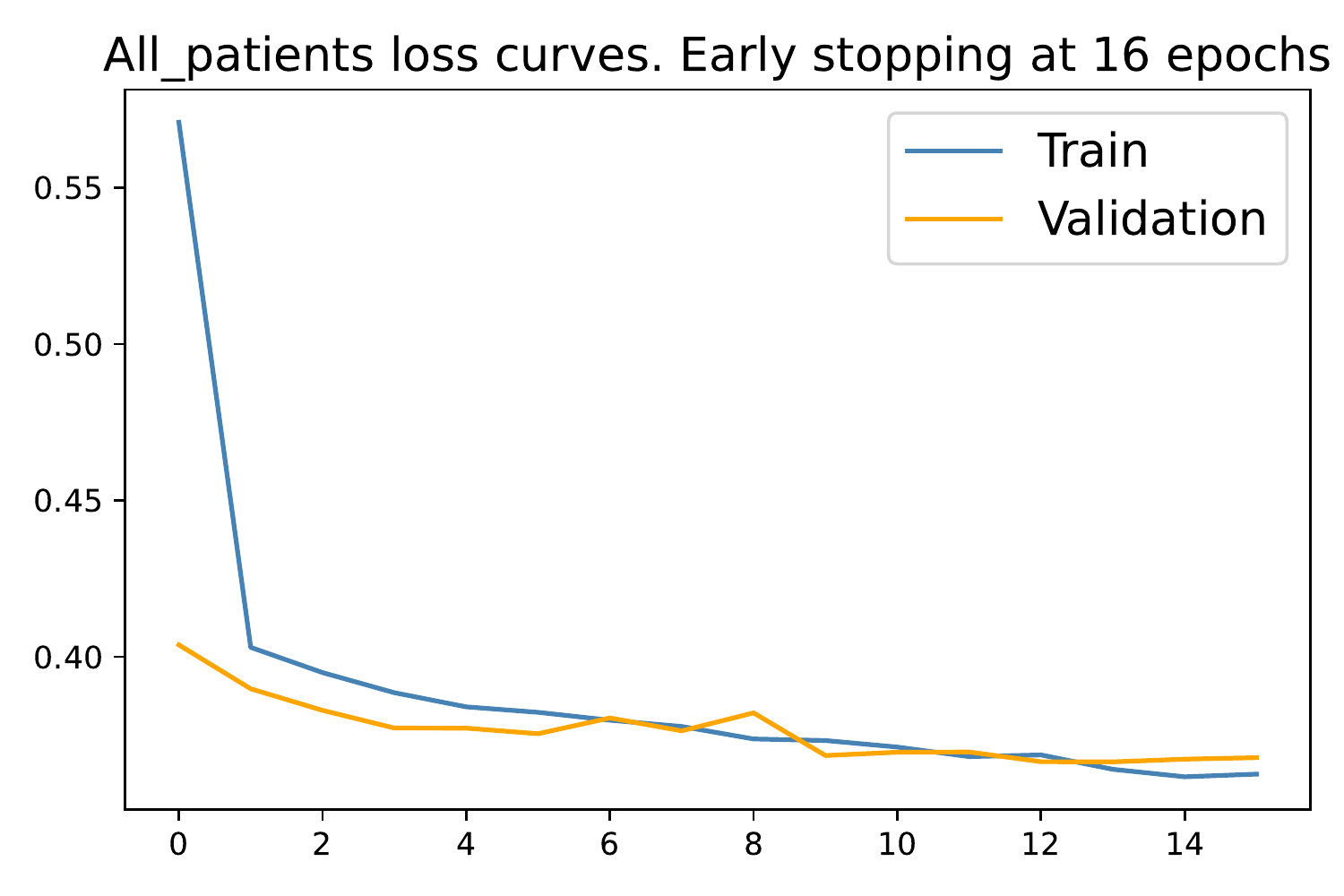}}
    \subfigure[Med-Surg ICU]{
\label{fig:med_surg_loss_mimic}
        \includegraphics[width=.27\linewidth]{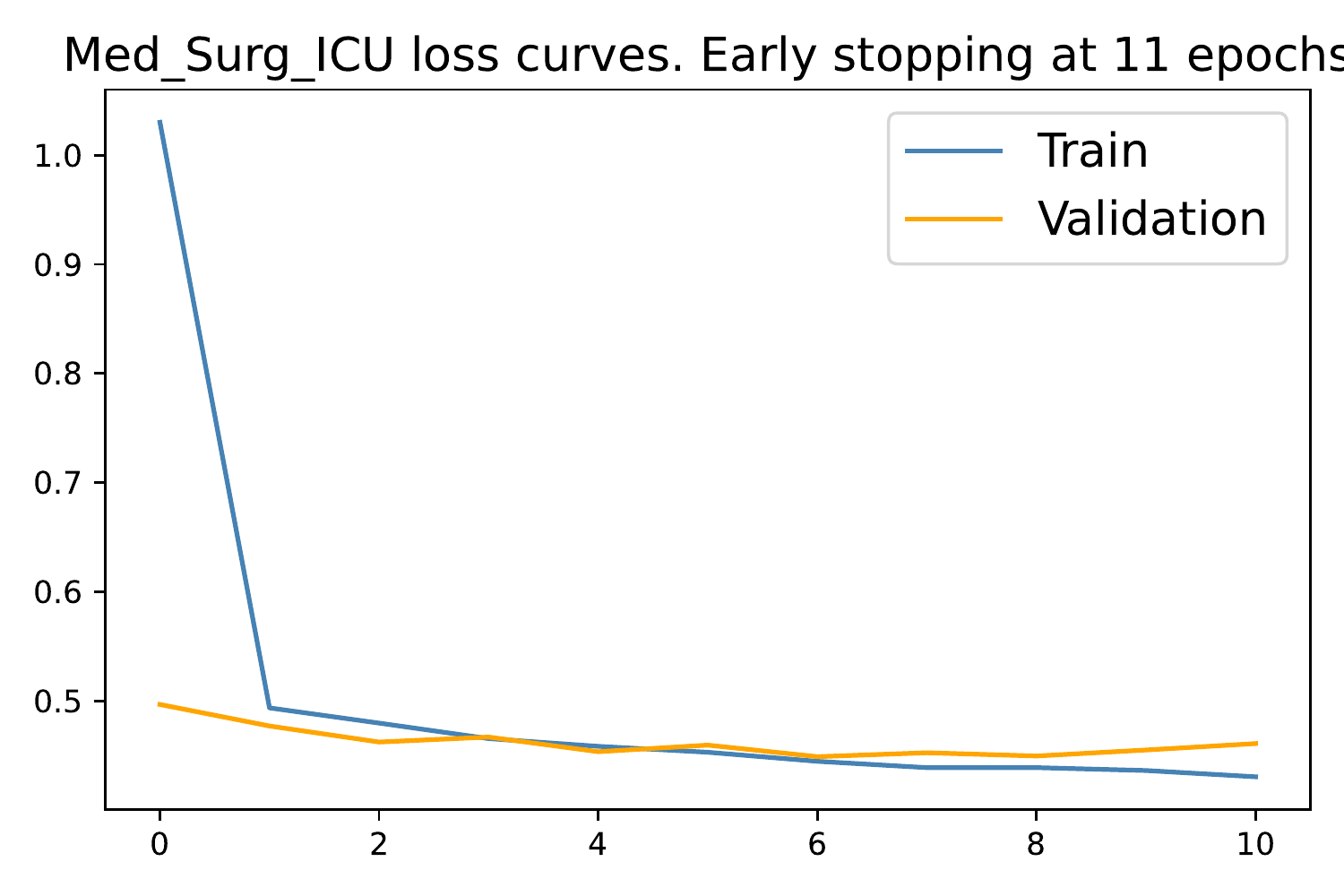}}
    \subfigure[TSICU]{
\label{fig:cticu_loss_mimic}
        \includegraphics[width=.27\linewidth]{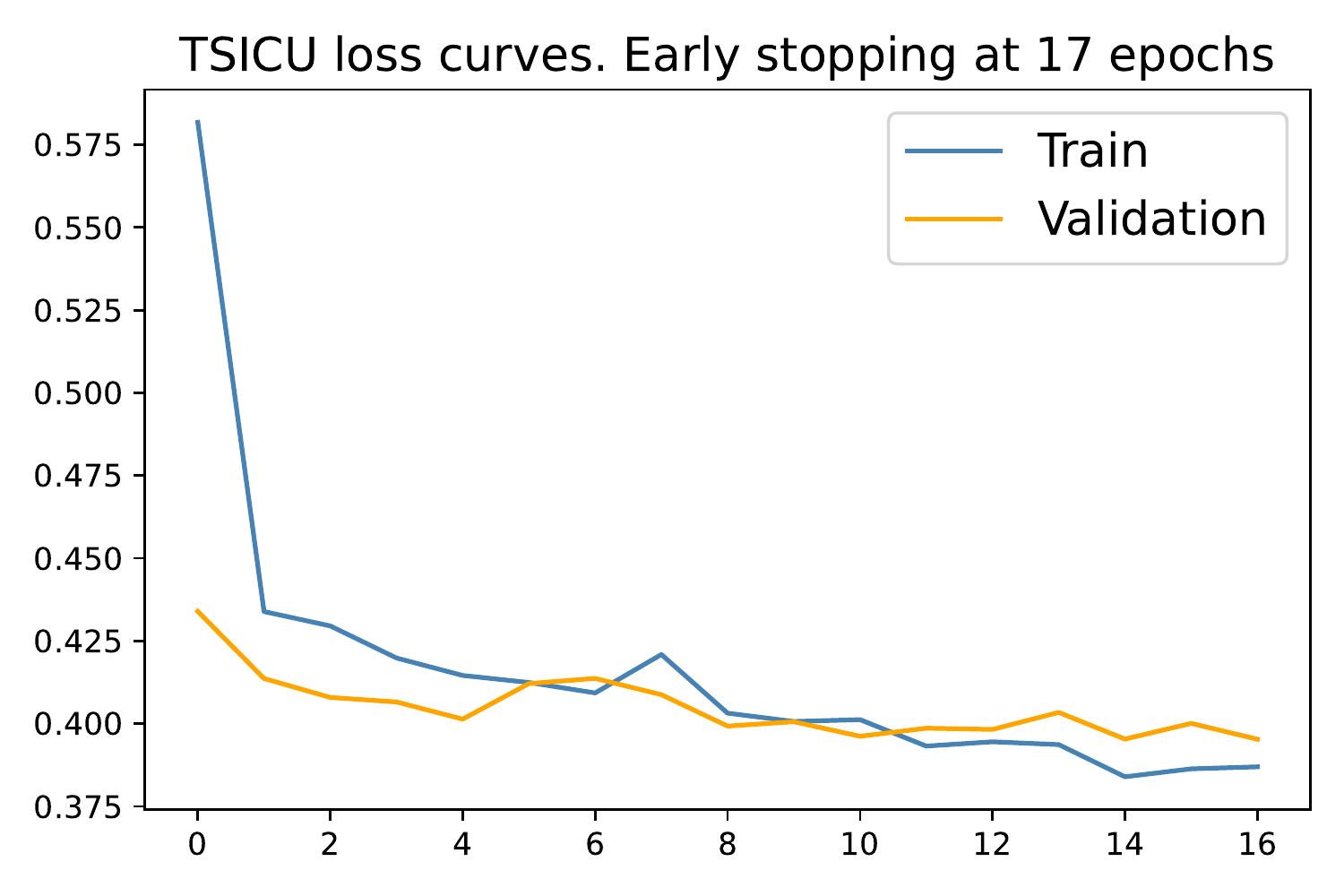}}
        \caption{Training and Validation curves obtained after model optimization on some MIMIC-IV units.}
    \label{fig:mimic_loss_curves}
\end{figure}

Training and model optimization were done following a systematic procedure (see Appendix \ref{bayes_hps_tune}) by carefully monitoring both the generalization error (error on test set) and the loss curves in Figures \ref{fig:eicu_loss_curves} and \ref{fig:mimic_loss_curves}. The batch size, the dropout, the number of hidden layers (No.layers), the number of hidden units (No.units) and the learning rate (lr) were all model hyperparameters obtained via Bayesian optimization \cite{o2019keras, omalley2019kerastuner} and are shown in Tables \ref{tab:eicu_hps} and \ref{tab:mimic_hps}. The Adam optimizer was used.
\subsubsection{eICU-CRD data}
\begin{table}[H]
    \caption{Hyperparamters results on eICU. S: source domain and T: target domain.}
    \centering
    \begin{tabular}{ccccccc}
    \toprule
    ICU data & batch size & No.layers & No.units & lr & dropout & domain type\\
    \midrule
      Med-Surg ICU & 512 &1 & 64 & 1e-3 & 0.1&S\\
      MICU & 128 & 1 & 32 & 8.99e-4 & 0.1 & T\\
      CCU-CTICU & 128 & 1 & 16 & 1.129e-3 & 0.2 & T\\
      NICU & 128 & 1 & 16 & 1e-3 & 0.2 & T\\
      CICU & 32 &1 &64 &1e-3 &0.2 & T \\
      SICU & 64 &1 & 32 & 8.99e-4 &0.2 & T \\
      CTICU & 64 &1 &16 & 1e-3 & 0.2  & T \\
      CSICU & 64 &1 & 64 &1e-4 & 0 & T \\
     \hline
   All stays & 512 & 1 & 64 & 0 & 1.129e-3 & \\
      \bottomrule
    \end{tabular}
    \label{tab:eicu_hps}
\end{table}

\subsubsection{MIMIC-IV data}
\begin{table}[H]
    \caption{Hyperparamters results on MIMIC-IV. S: source domain and T: target domain.}
    \centering
    \begin{tabular}{ccccccc}
    \toprule
    ICU data & batch size & No.layers & No.units & lr & dropout & domain type\\
    \midrule
      MICU & 128 & 1 & 32 & 7.142e-4 & 0.3 & S or T\\
      CVICU & 128 & 1 & 64 & 5.46e-3 & 0.3 & S or T\\
      Med-Surg ICU & 128 & 1 & 64 & 8.99e-4 & 0.2& S or T\\
      SICU & 128 &1 & 64 & 4.833e-4 &0.2 & S or T \\
      TSICU & 64 &1 & 64 & 1e-2 & 0.3 & T \\
      CCU & 64 & 1 & 16 & 1e-2 & 0.1 & T\\
      Neuro SICU & 16 & 1 & 8 & 6.952e-4 & 0.2 & T\\
      NI & 16 &1 & 16 & 4.833e-4 & 0.2 & T \\
      NS & 8 &1 &16 &1.624e-4 &0.1 & T \\
     \hline
    All stays & 512 & 1 & 64 & 0.1 & 1.438e-3 & \\
      \bottomrule
    \end{tabular}
    \label{tab:mimic_hps}
\end{table}

\subsection{Baseline and weight transfer models}
\label{sec:results}

The performance of the weight transfer model was compared to that of the baseline model, where individual models were constructed for each unit using the hyperparameters obtained from optimization and listed in Tables \ref{tab:eicu_hps} and \ref{tab:mimic_hps}. Also, for comparisons with the common practice in the literature,
a model was ran where all units are considered as a single homogeneous entity and their data learned together in a single ``all stay" model. For the weight transfer model, all hyperparameters except the batch size from the source are transferred to the targets. All models were ran 100 times by shuffling the data to obtain a distribution of the prediction error on the test set. 95\% confidence intervals were estimated by calculating the $2.5$th and $97.5$th percentiles of the prediction error on the test set. Pairwise t-test in population means between ``all stay" model and each population model in Tables \ref{tab:eICU_baseline_results} and \ref{tab:mimic_baseline_results} are indicated with stars (* p$<0.05$, **p$<0.001$). The red stars on boxplot figures indicate instances where the model stagnates, i.e., doesn't learn and stops after a few epochs due to the early stopping condition (that prevents overfitting).
\subsubsection{eICU-CRD data}
\begin{figure}[H]
    \centering
    \subfigure[Number of epochs to reach convergence.]{
 \label{fig:eicu_epochs}
         \includegraphics[width=.40\linewidth]{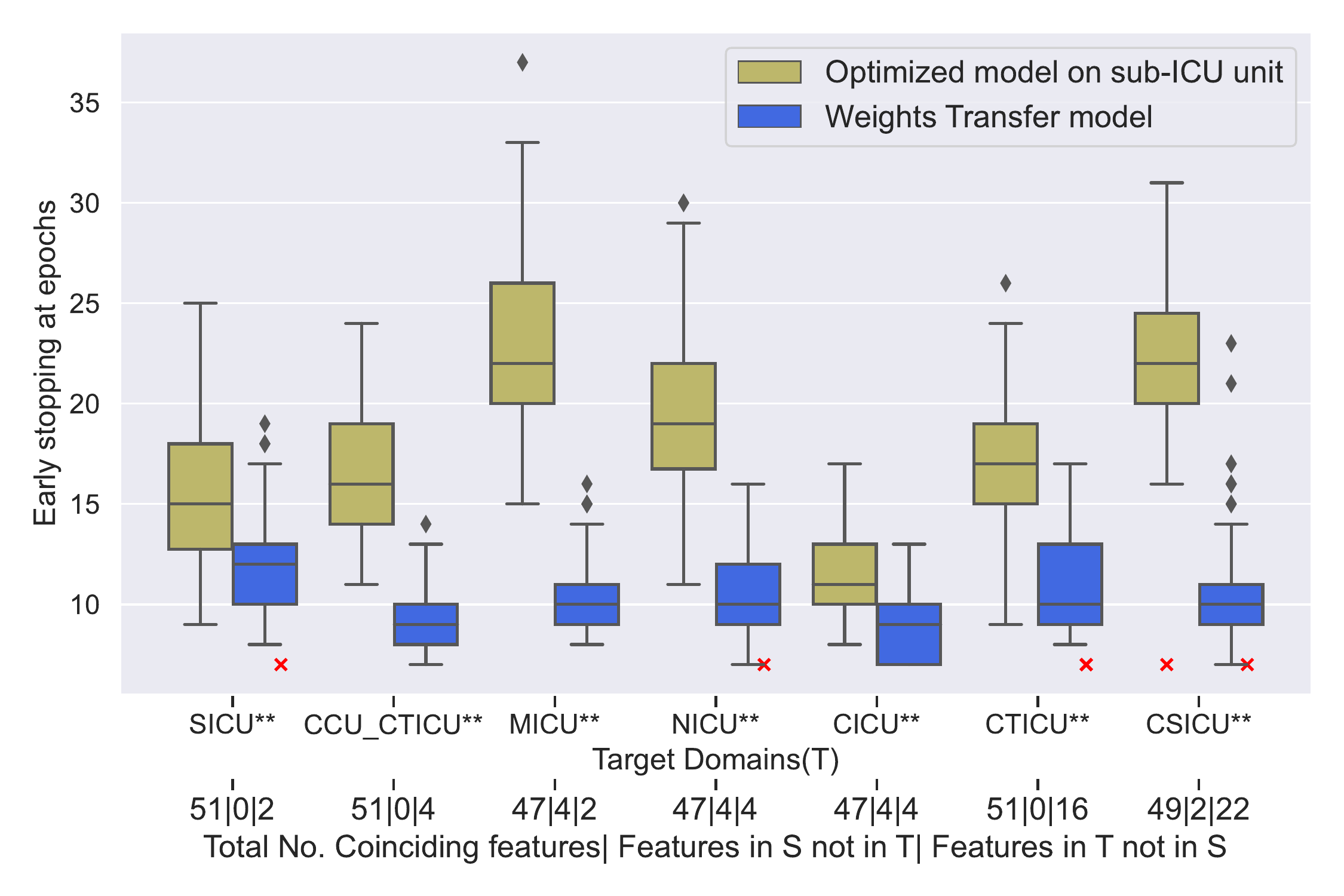}}
         \hspace{1.0cm}
    \subfigure[Mean absolute error in days on test set]{
\label{fig:eicu_error}
        \includegraphics[width=.40\linewidth]{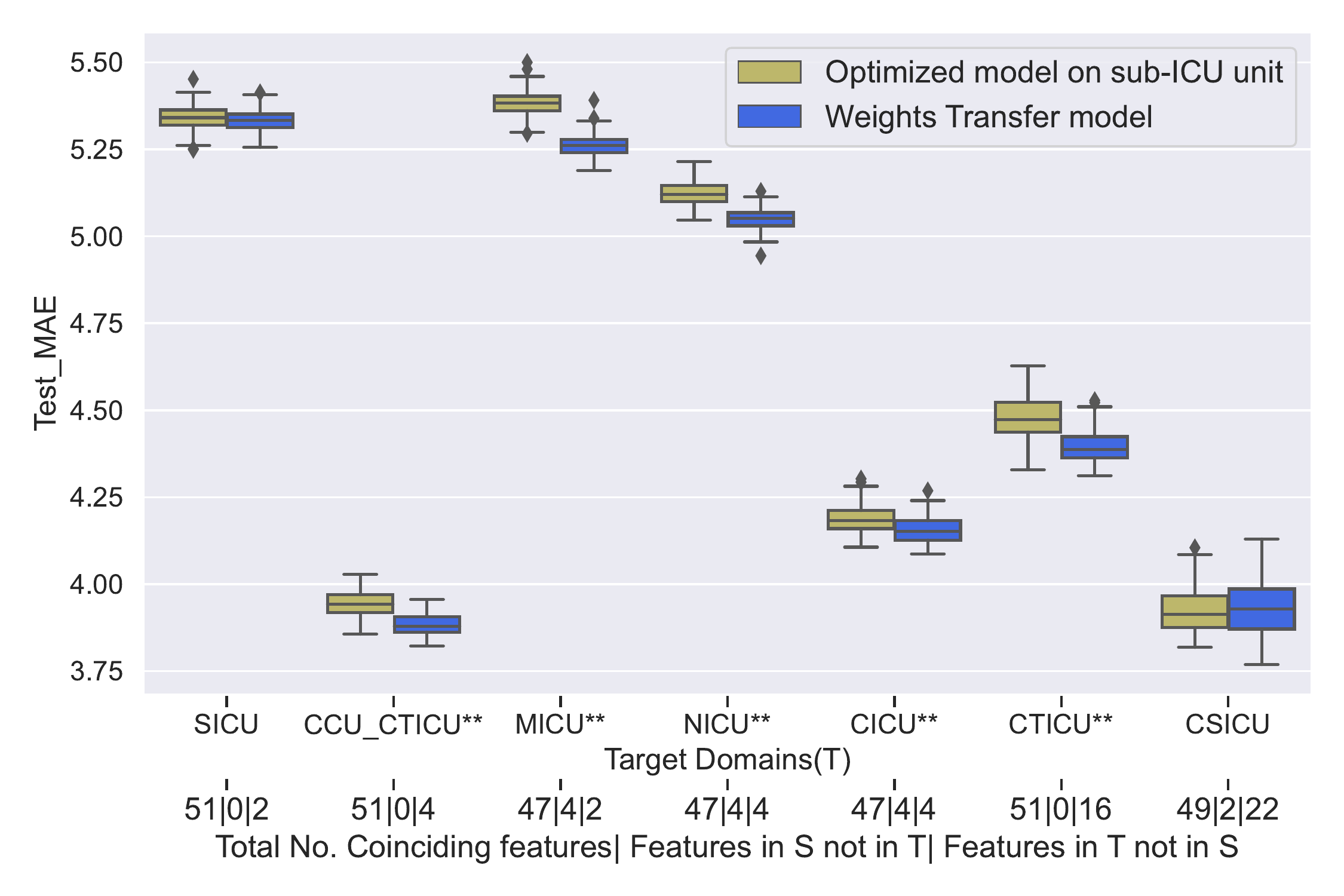}}
        \caption{Distribution of number of epochs and error measures on Test set with (blue) and without (lemon green) transfer learning on eICU. Statistically significance per unit performed using a t-test (* p$<0.05$, **p$<0.001$).}
    \label{fig:eicu_results}
\end{figure}

Table \ref{tab:eICU_baseline_results} shows that feeding the model with all ICU stays data for LoS predictions irrespective of their admission units can sometimes overestimate the error in a unit. Looking at the most interpretable error measure, that is, MAE (reported in fractional days), we see that the error of the ``all stay" model significantly over estimates that of Med-Surg ICU, CCU-CTICU, CICU and CSICU by 3.9, 6.35, 1 and 8.69 hours respectively.
\begin{table}[H]
    \caption{eICU-CRD: Baseline models for each unit and ``all stay model": error estimates on test set with 95\% confidence intervals: All stays data fitted in one model and one model per ICU domain (individuals models are learned and trained on each ICU domain).}
    \begin{tabular}{p{2.5cm}p{3.8cm}p{3.8cm}p{4.2cm}}
    \toprule
    ICU unit & MAE & MAPE & MSE \\
    \midrule
    Med-Surg ICU & 4.0311(4.0183-4.0559)** & 0.6909(0.6614-0.7111)** & 45.2873(44.8557-46.2811)**\\
    MICU & 5.3410(5.3158-5.4591)** &0.7722(0.7692-0.8431)**&80.4968(77.6054-82.2602)**\\
    CCU-CTICU & 3.9303(3.8816-4.0206)** & 0.6497(0.6532-0.7227)** &49.6607(47.1492-50.0401)**\\
    NICU & 5.1263(5.0596-5.2009)** & 0.8481(0.7985-0.8689)** &68.9331(67.0461-70.8095)**\\
    CICU & 4.1531(4.1224-4.2743)**& 0.7305(0.6959-0.8269)**& 48.8806(46.7863-50.0562)**\\
    SICU & 5.2859(5.2696-5.2966)** &0.6854(0.6463-0.7260)** &78.4585(77.5238-83.1262)**\\
    CTICU &  4.5133(4.3767-4.5937)** &0.5555(0.5179-0.5918)** &59.7552(55.8821-60.9689)**\\
    CSICU & 3.8328(3.8235-4.0838)**& 0.6930(0.6703-0.7971)** &41.4605(41.4058-47.2725)**\\
    \hline
    All stays & 4.1948(4.1851-4.2203) &0.6786(0.6568-0.7030)& 50.3304(49.9702-51.4429)\\
    \bottomrule
    \end{tabular}
    \label{tab:eICU_baseline_results}
\end{table}
\noindent Figure \ref{fig:eicu_results} shows the effect of using pre-trained weights from $S$ as initial weights for coinciding features and assigning all hyperparameters (except the batch size) from $S$ to $T$. In Figure \ref{fig:eicu_epochs}, we see that model convergence always occurs significantly faster even when a good number of the features in $T$ are not in $S$, as it is the case for CSICU and CTICU. Figure \ref{fig:eicu_error} shows that for most of the units, specifically, CCU-CTICU, MICU, NICU, CICU and CTICU, weight transfer significantly improves prediction accuracy.
\subsubsection{MIMIC-IV data}
\begin{figure}[H]
    \centering
    \subfigure[Source domain (S): Med-Surg ICU]{
 \label{fig:med_surg_source_error}
         \includegraphics[width=.40\linewidth]{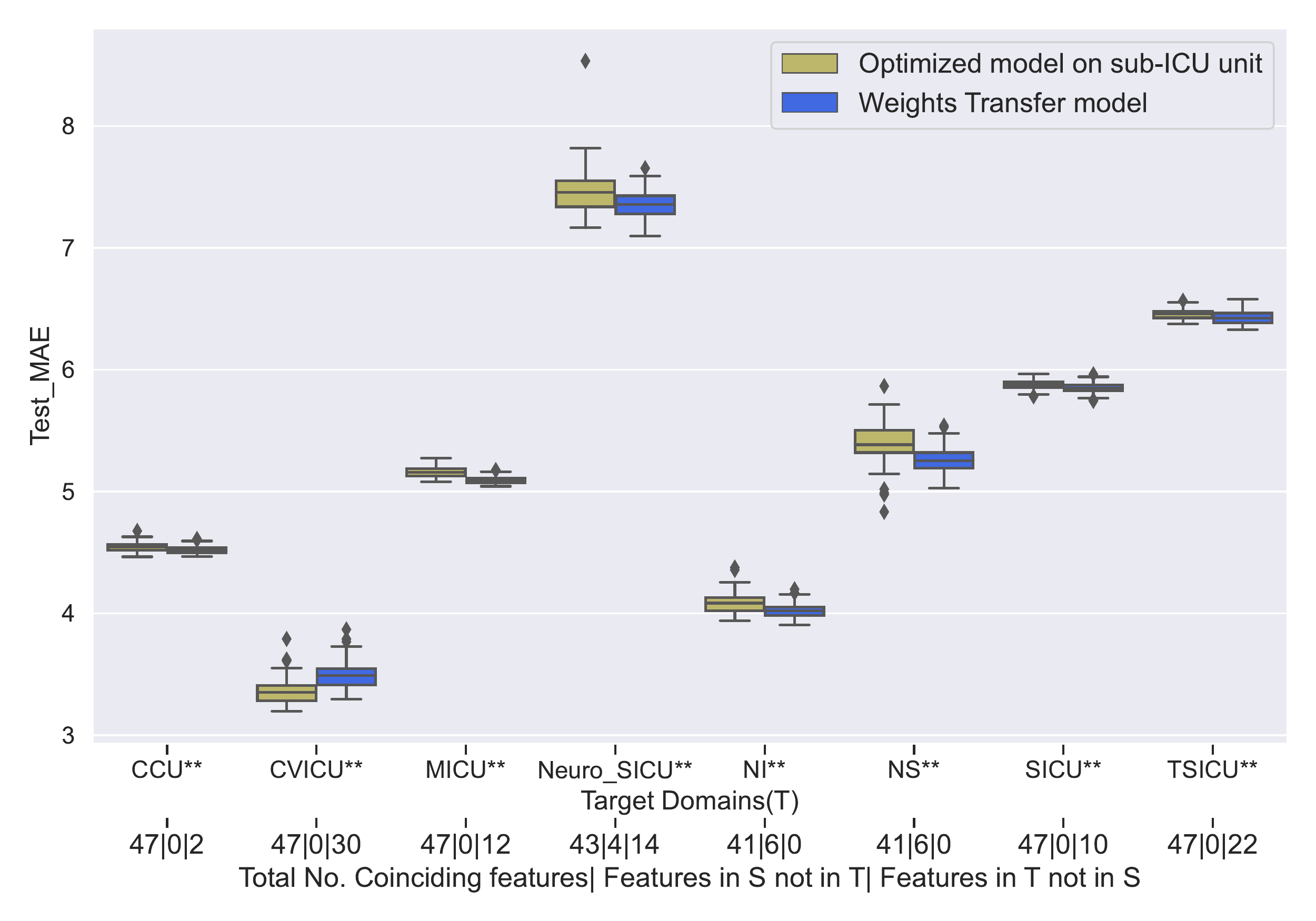}}
         \hspace{1.0cm}
    \subfigure[Source domain (S): MICU]{
\label{fig:micu_source_error}
        \includegraphics[width=.40\linewidth]{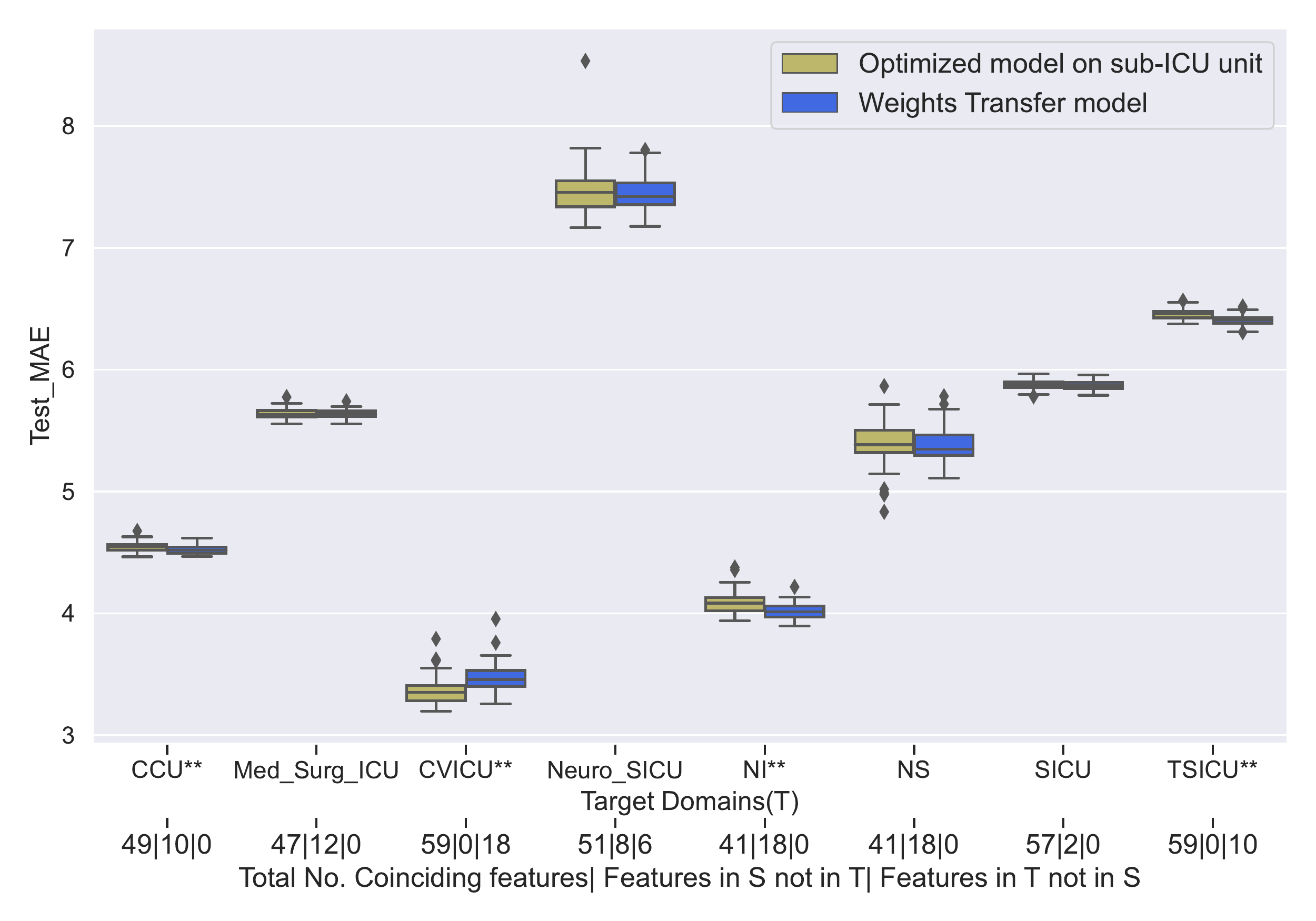}}
        \hspace{1.0cm}
    \subfigure[Source domain (S): SICU]{
\label{fig:sicu_domain_error}
        \includegraphics[width=.40\linewidth]{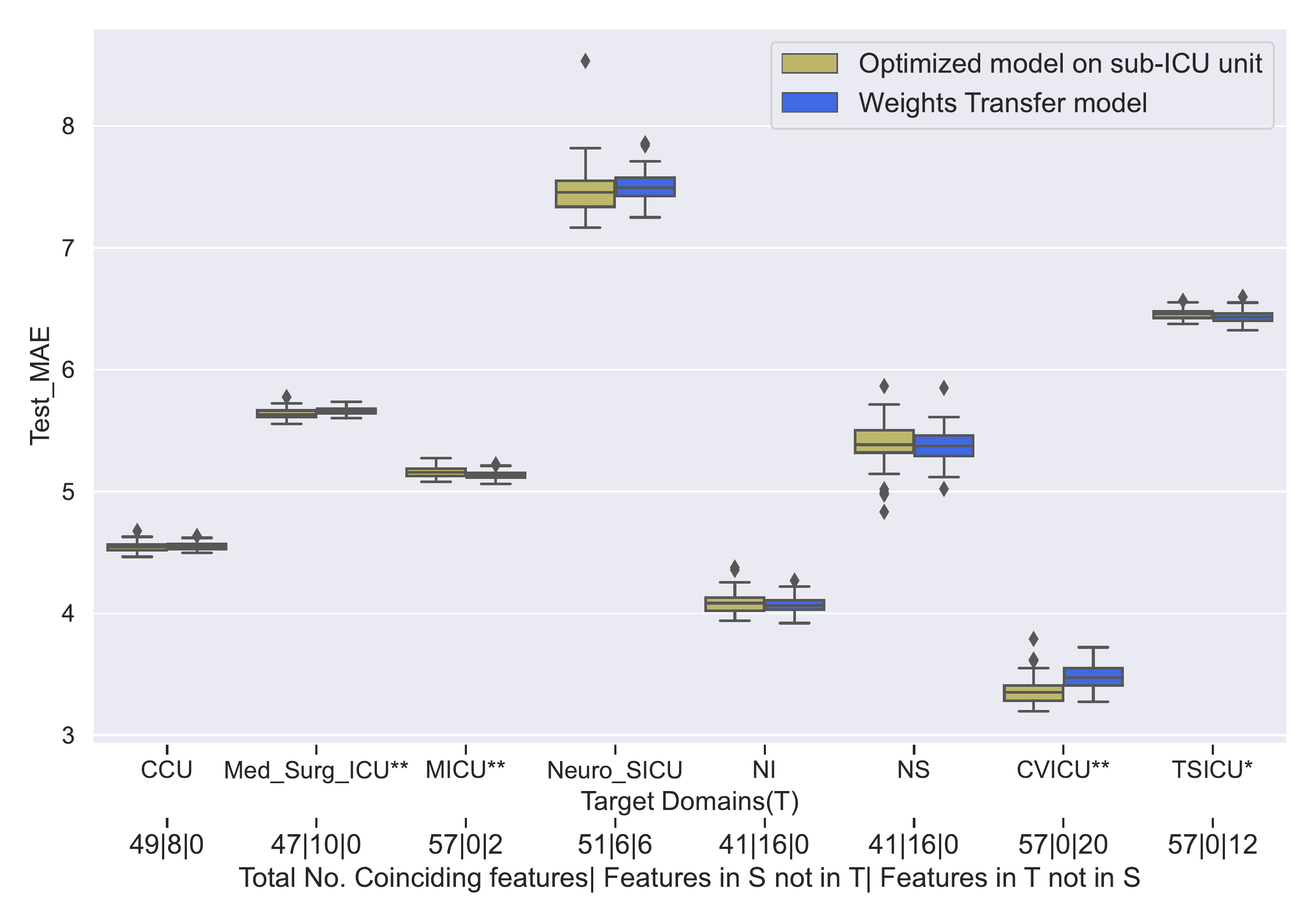}}
        \hspace{1.0cm}
    \subfigure[Source domain (S): CVICU]{
\label{fig:cvicu_domain_error}
        \includegraphics[width=.40\linewidth]{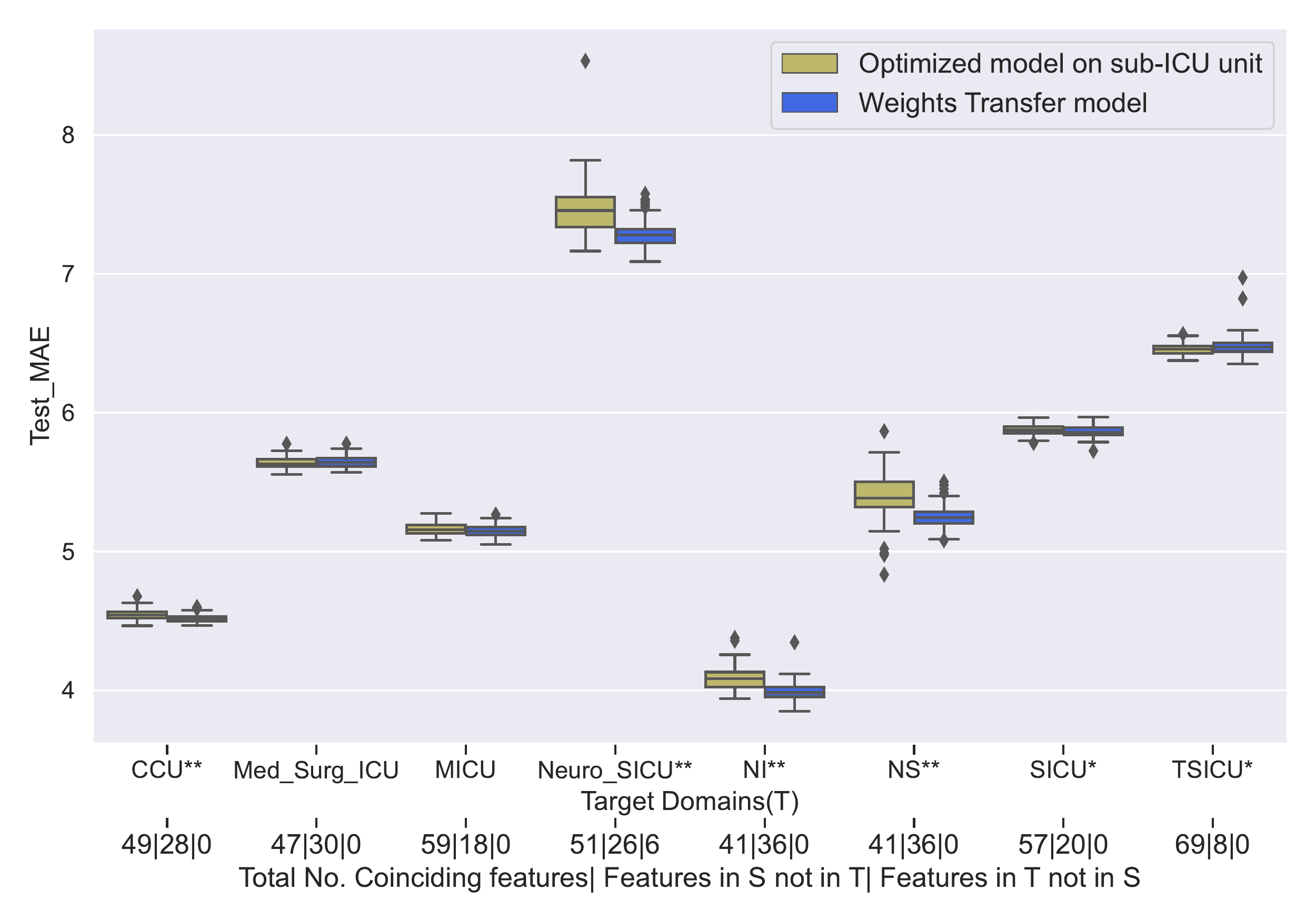}}
        \caption{Distribution of MAE error on test set with (blue) and without (lemon green) transfer learning on MIMIC-IV with Med-Surg ICU, MICU, SICU and CVICU as potential source domains. Statistically significance per unit performed using a t-test (* p$<0.05$, **p$<0.001$).}
    \label{fig:mimic_error_results}
\end{figure}

The observation drawn from Table \ref{tab:eICU_baseline_results} holds for MIMIC-IV data in Table \ref{tab:mimic_baseline_results}, where the ``all stay" model does not always perform significantly better than unit models as seen for CVICU, CCU and NI. For these units, the ``all stay" model MAE error over estimates the MAE error by 1.79, 0.57 and 1.15 days respectively.
\begin{table}[H]
    \caption{MIMIC-IV: Baseline models for each unit and ``all stay" model: error estimates on test set with 95\% confidence intervals: All stays data fitted in one model and one model per ICU domain.} 
    \centering
    \begin{tabular}{p{2.5cm}p{3.8cm}p{3.8cm}p{4.2cm}}
    \toprule
    ICU unit & MAE & MAPE & MSE \\
    \midrule
    Med-Surg ICU & 5.6835(5.5664-5.7218)** & 0.8407(0.7480-0.8633)** & 91.4511(88.7403-94.8977)**\\
    MICU & 5.1596(5.0864-5.2703) & 0.7046(0.6942-0.7968)** & 68.3557(65.0615-69.3762)**\\
    CVICU & 3.3207(3.2265-5.9338)** & 0.4041(0.3746-0.7682)** & 42.7216(40.8025-78.0702)**\\
    SICU & 5.8401(5.8017-5.9516)** & 0.6782(0.6808-0.7909)** & 92.8705(87.6492-93.3682)**\\
    TSICU &  6.4758(6.3811-6.5416) & 0.7936(0.6962-0.8180) & 102.7655(98.9423-108.4076)\\
    CCU & 4.5474(4.4818-4.6177)** & 0.7174(0.6637-0.7590)** & 62.6330(61.2112-64.7852)** \\
    Neuro SICU & 7.2245(7.1893-8.1933)** & 0.9108(0.8435-1.0569)** & 141.0862(132.4134-220.3568)**\\
    NI & 3.9619(3.9564-4.2511)** & 0.6184(0.6421-0.7700)** & 50.7301(48.9124-55.6811)**\\
    NS & 5.5758(5.0034-5.7128)** & 0.8699(0.6902-0.9102)** & 94.2374(83.9943-102.3467)**\\
    \hline
    All stays & 5.1140(5.0508-5.1424) & 0.6687(0.6125-0.6781) & 73.8215(73.0283-76.9459)\\
    \bottomrule
    \end{tabular}
    \label{tab:mimic_baseline_results}
\end{table}
\begin{figure}[H]
    \centering
    \subfigure[Source domain (S): Med-Surg ICU]{
 \label{fig:med_surg_source_mimic}
         \includegraphics[width=.40\linewidth]{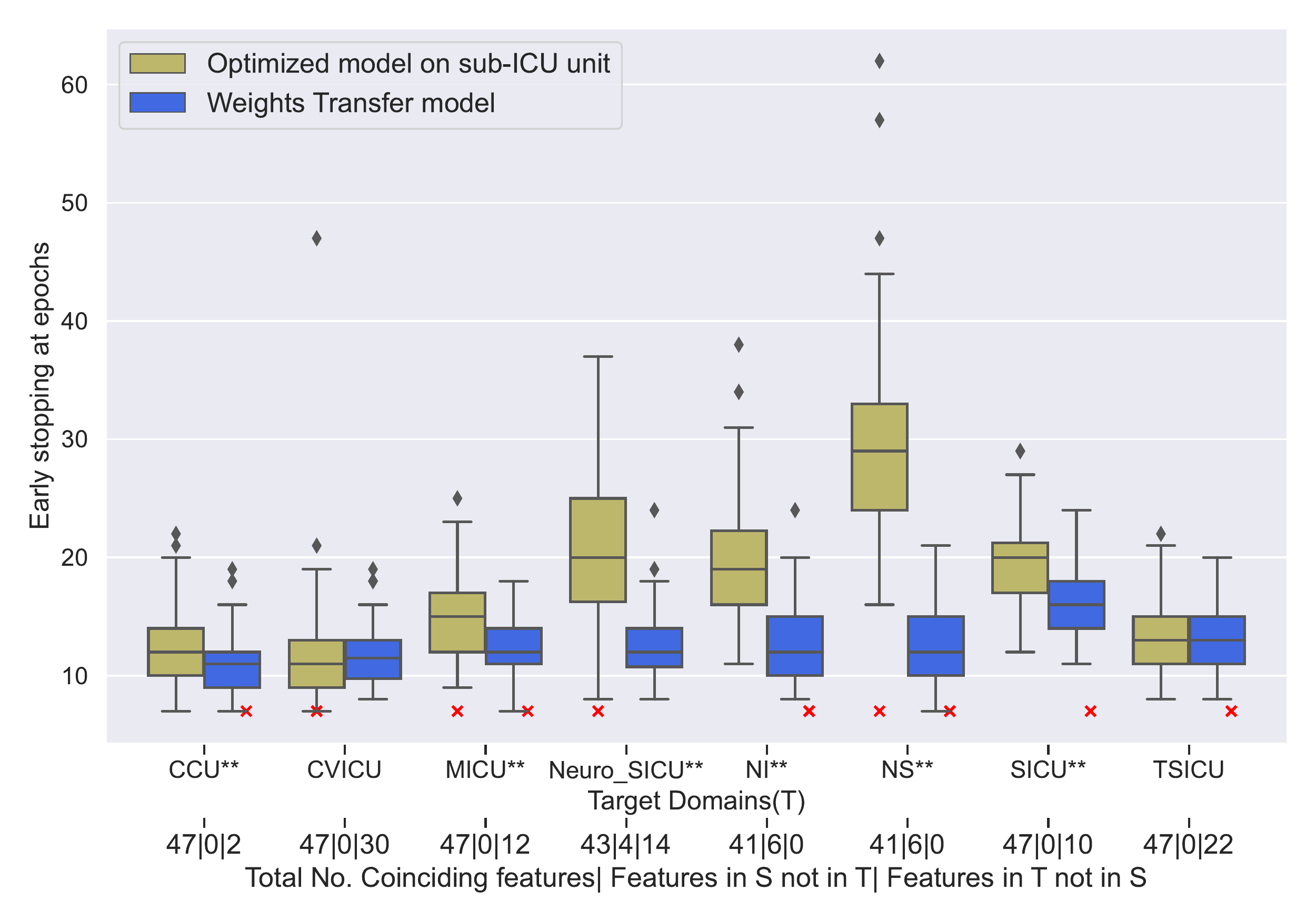}}
         \hspace{1.0cm}
    \subfigure[Source domain (S): MICU]{
\label{fig:micu_source}
        \includegraphics[width=.40\linewidth]{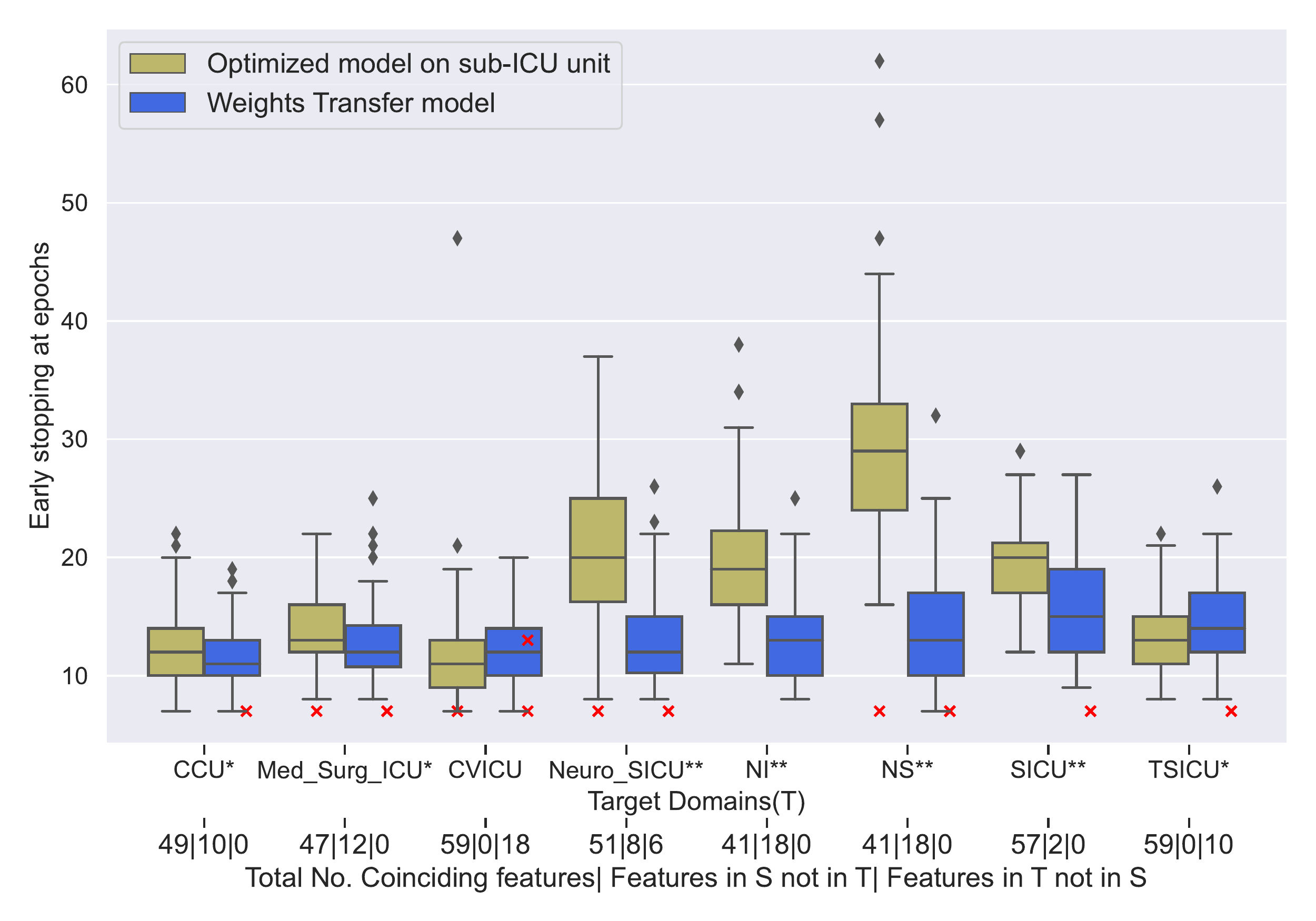}}
        \hspace{1.0cm}
    \subfigure[Source domain (S): SICU]{
\label{fig:sicu_domain}
        \includegraphics[width=.40\linewidth]{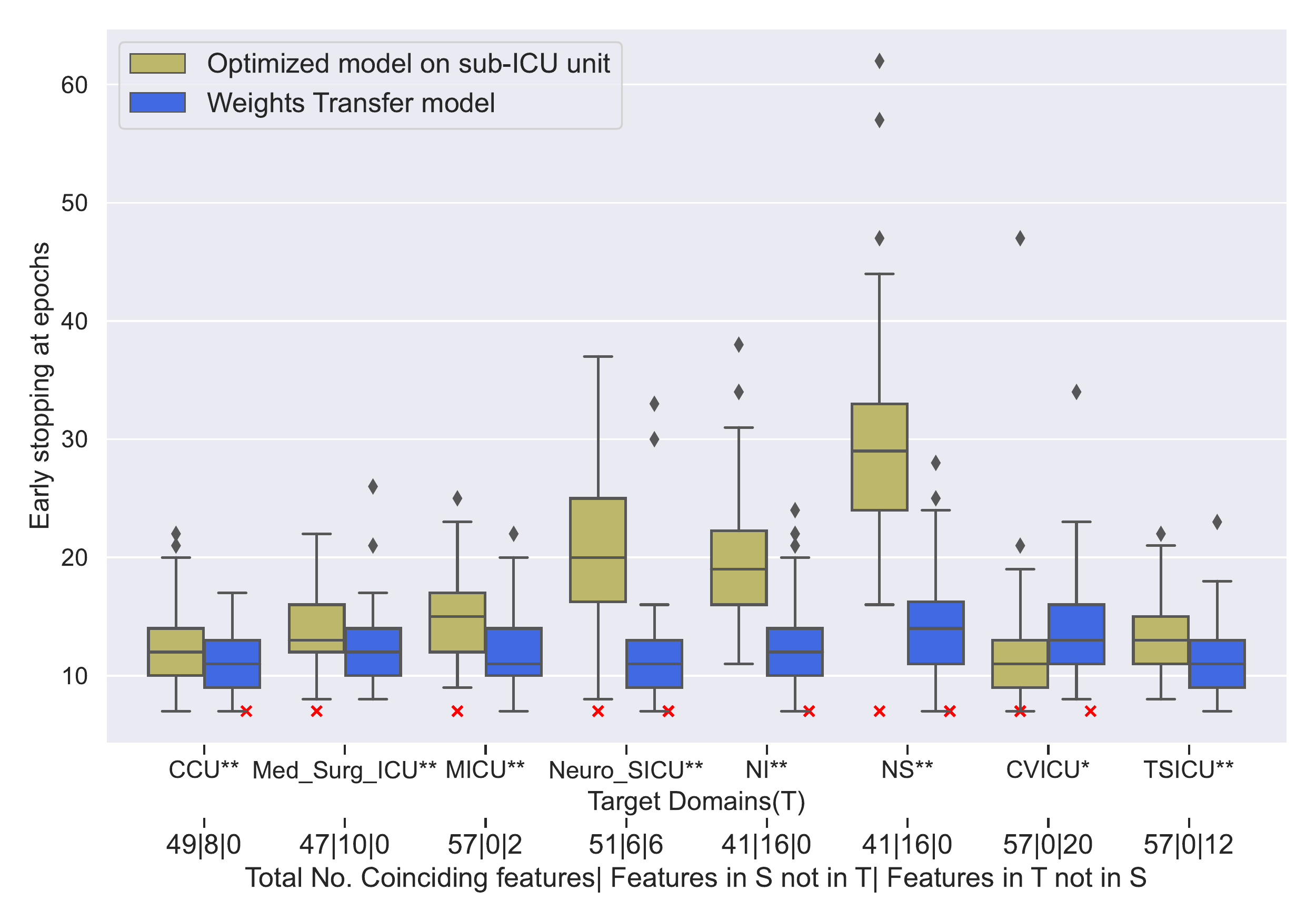}}
        \hspace{1.0cm}
    \subfigure[Source domain (S): CVICU]{
\label{fig:cvicu_domain}
        \includegraphics[width=.40\linewidth]{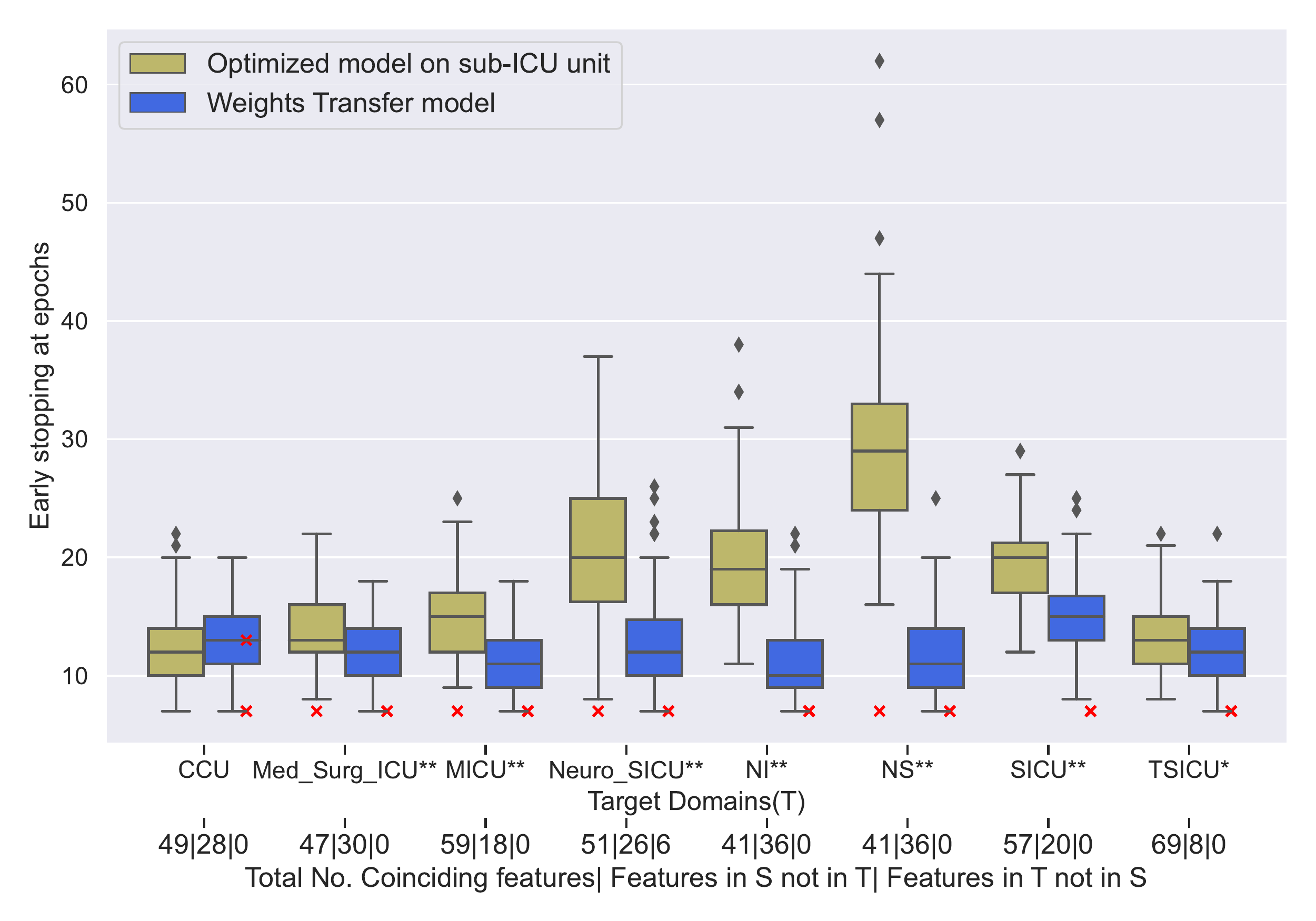}}
        \caption{Distribution of number of epochs to reach convergence with (blue) and without (lemon green) transfer learning on MIMIC-IV with Med-Surg ICU, MICU, SICU and CVICU as potential source domains. Statistically significance per unit performed using a t-test (* p$<0.05$, **p$<0.001$).} 
    \label{fig:mimic_epochs_results}
\end{figure}
Looking at the four potential source domains on Figures \ref{fig:mimic_epochs_results} and \ref{fig:mimic_error_results}, low populated ICUs like Neuro-SICU, NI and NS experience the highest gains with all source domains. As target domain, Med-Surg ICU gives the overall best improvement either in computation time, prediction accuracy or even both (see Figures \ref{fig:med_surg_source_mimic} and \ref{fig:med_surg_source_error}). As target domain, CVICU is negatively impacted by weight transfer (Figures \ref{fig:cvicu_domain} and \ref{fig:cvicu_domain_error}).
\subsection{Model Interpretability}
\begin{figure}[H]
    \centering
    \subfigure[CTICU patients]{
\label{fig:cticu_xai}
        \includegraphics[width=.3\linewidth]{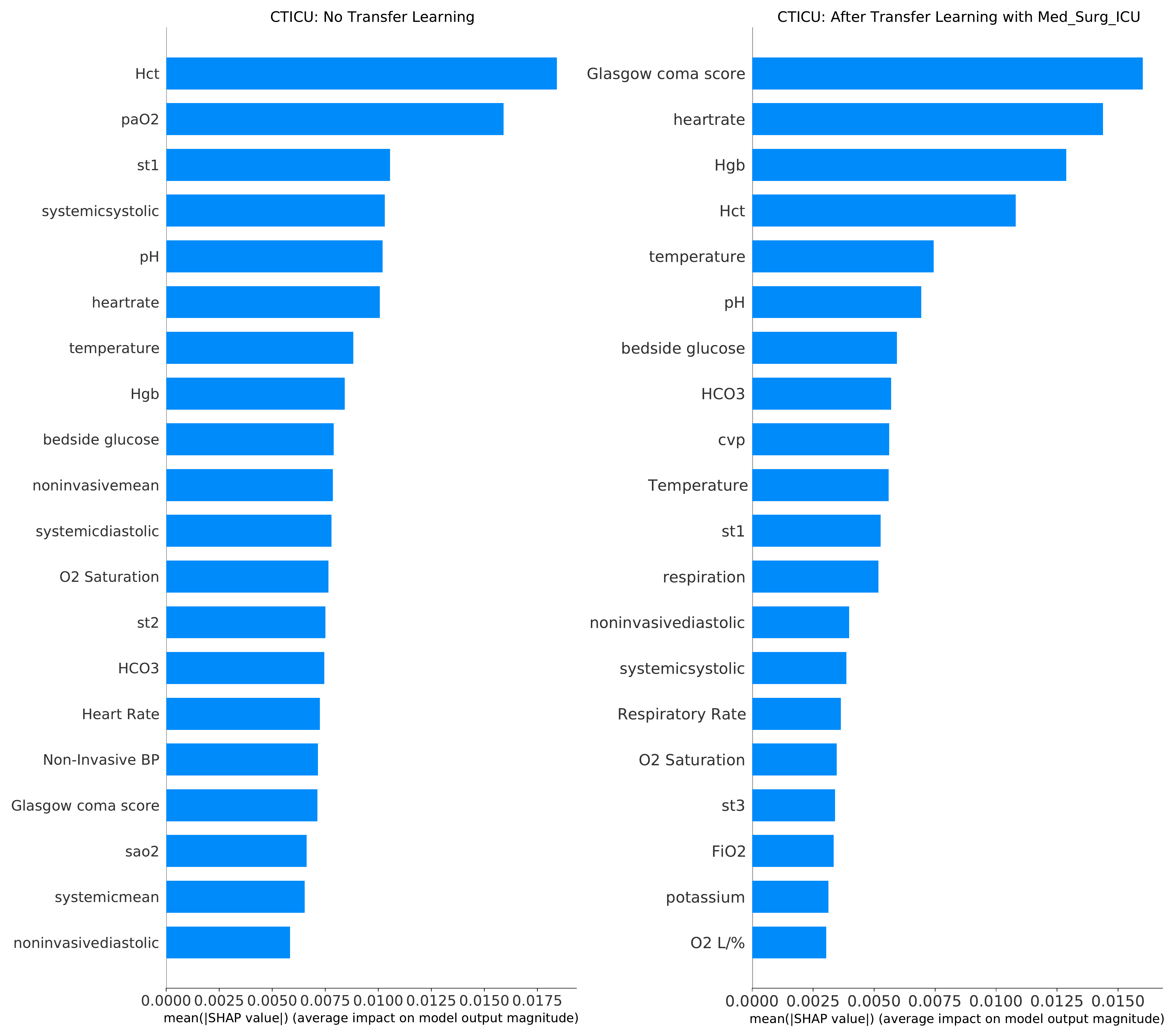}}
 \hspace{1.0cm}        
\subfigure[TSICU patients]{
\label{fig:tsicu_xai}
\includegraphics[width=.3\linewidth]{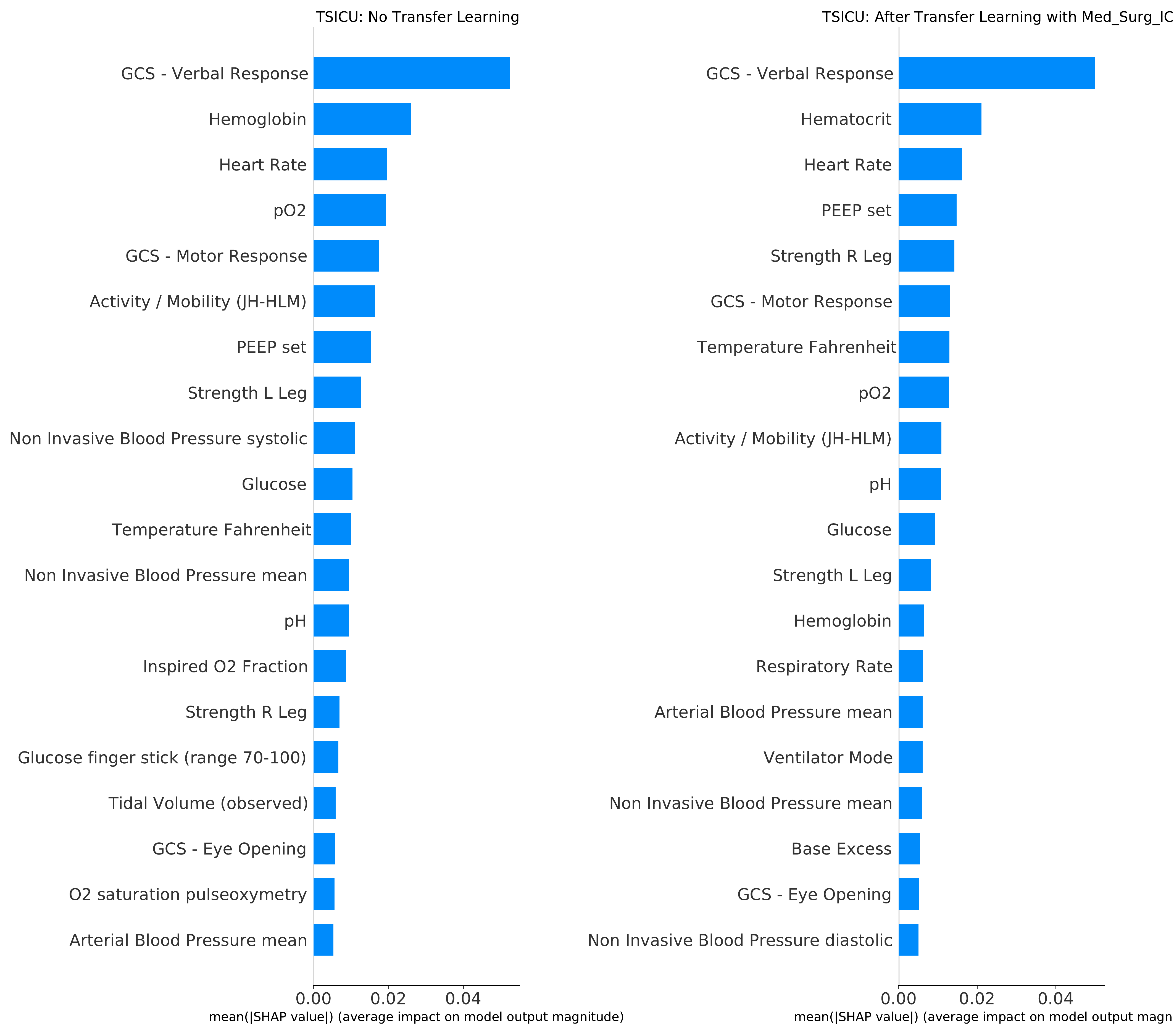}}
\caption{Top 25 most important features on CTICU (from eICU) and TSICU (from MIMIC-IV) targets before and after applying weight transfer}
\label{fig:eicu_SICU_CTICU_xai}
\end{figure}

Given a trained model and the 3D training data set (number of patients $\times$ number of time-steps $\times$ number of inputs), 3D feature contributions for model prediction on the test set are obtained for each patient, at each time-step and for each input. Overall feature importance is then plotted by averaging over the time dimension and subsequently the patient dimension.\\
Patient features which can globally be split under charted parameters and labs are observed to be present before and after applying weight transfer as seen in Figure \ref{fig:eicu_SICU_CTICU_xai}, though the order is not conserved.

\subsection{Further Analyses}
To further understand the benefits of domain adaptation, we conducted three further analyses using eICU-CRD dataset. The first in which we do not optimize hyperparameters on each domain $T$ and use those found on $S$, as seen in Figure \ref{fig:eicu_results_hps_source}. This was performed in order to understand whether weight transfer gains were due to poor hyperparameter optimization on each target unit. In the second analysis, non-coinciding features between $S$ and $T$ were removed and complete model transfer (including pre-trained weights and model optimizer state) was carried out. As final analysis, different learning rates were imposed on input feature groups as explained in section \ref{sec:diff_lr} and Eq(\ref{eqn:diff_lr}) with $\alpha = 10^{-1}$.
\subsubsection{No hyperparameter optimization on target units}
\label{hps_source}
Here, hyperparameters on all lemon green boxplots are obtain from the trained source domain model.
\begin{figure}[H]
    \centering
    \subfigure[Epochs]{
 \label{fig:epochs_source}
         \includegraphics[width=.40\linewidth]{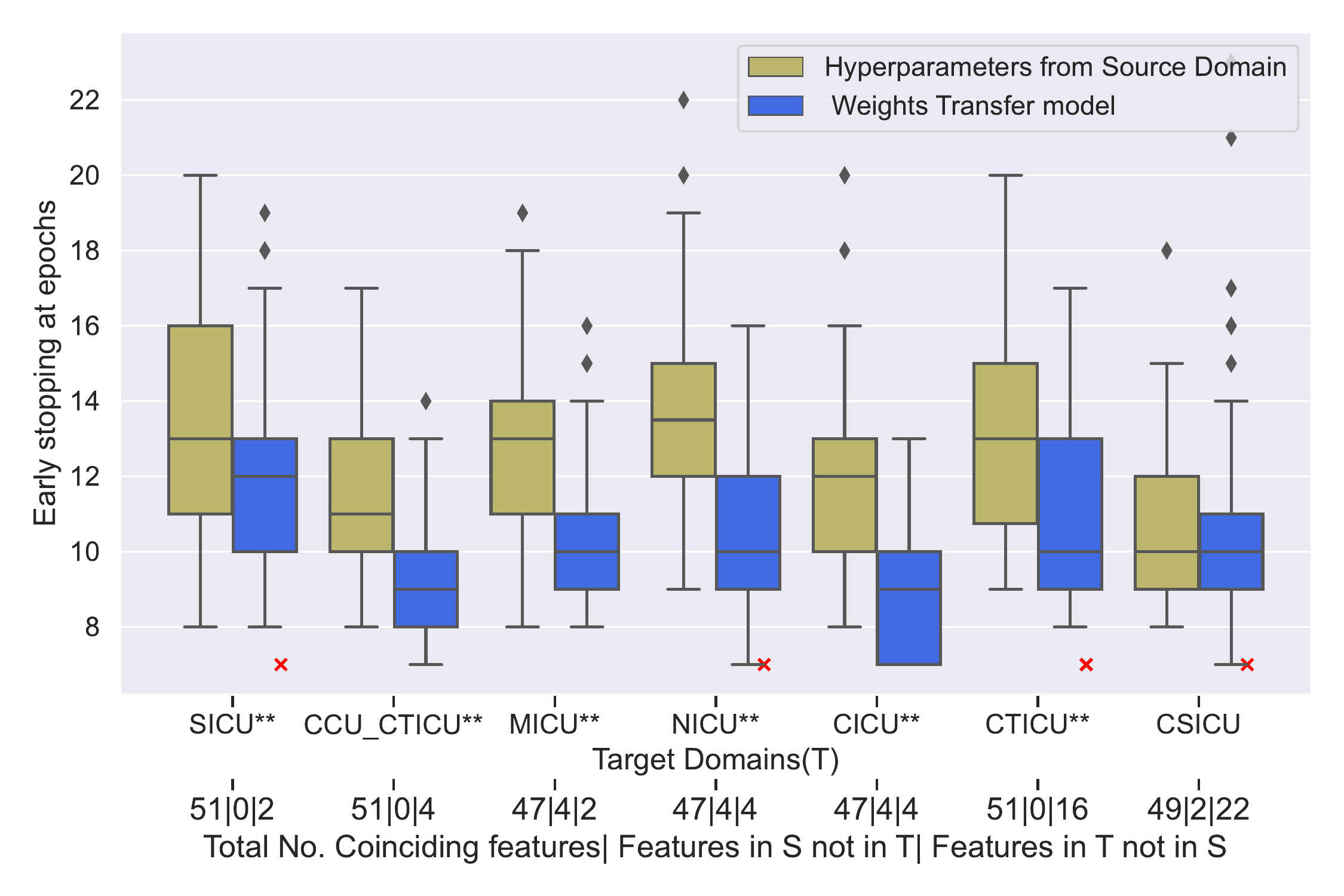}}
         \hspace{1.0cm}
    \subfigure[Error on test set]{
\label{fig:error_source}
        \includegraphics[width=.40\linewidth]{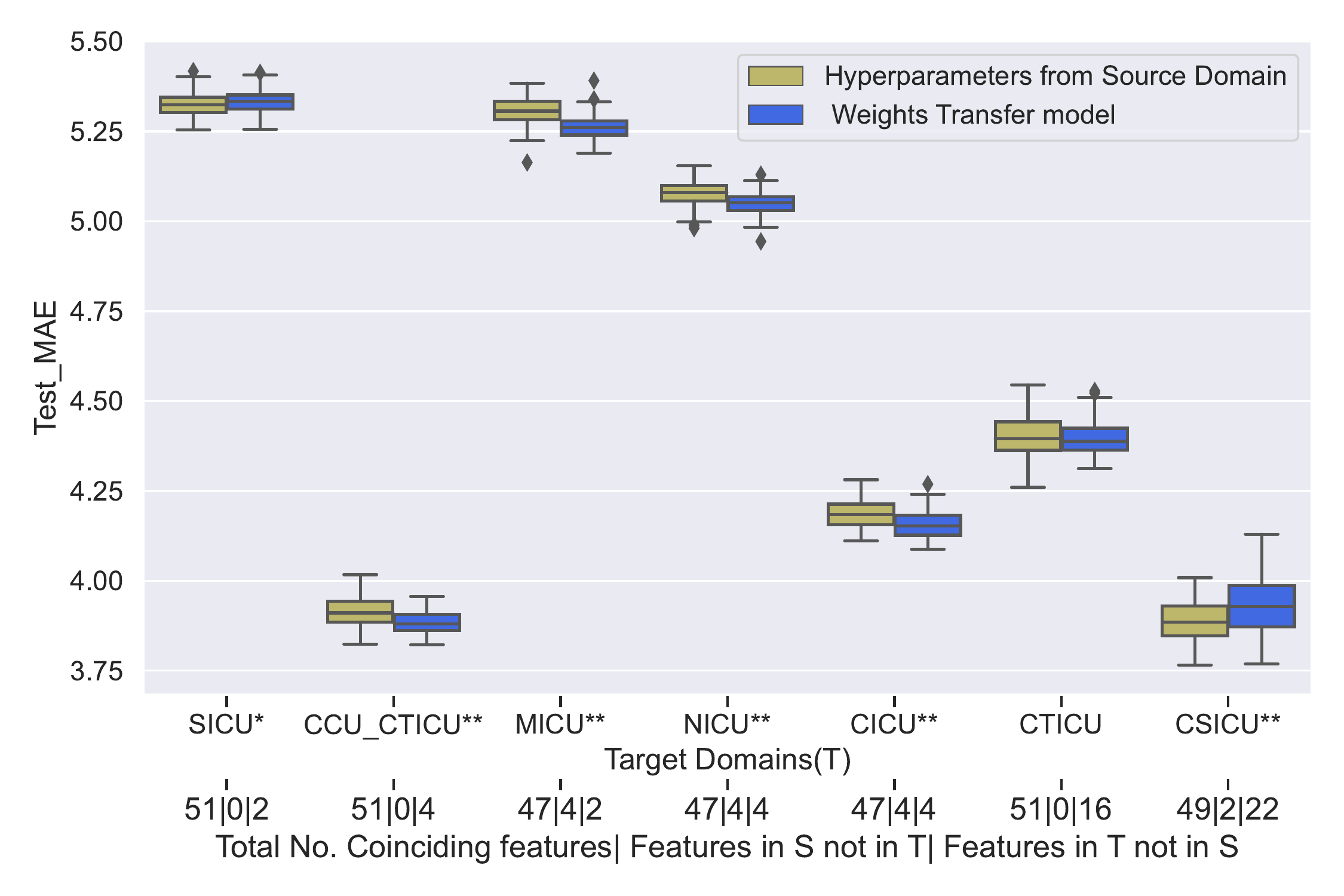}}
        \caption{Number of epochs and error measures with (blue) and without (lemon green) transfer learning on eICU-CRD where hyperparameters are fixed from source domain.}
    \label{fig:eicu_results_hps_source}
\end{figure}
Figures \ref{fig:epochs_source} and \ref{fig:error_source} show that pre-trained weights do speed model convergence and improve prediction accuracy even when both models (with and without weight transfer) have the same hyperparameters.
\subsubsection{Full Model Transfer}
Here non-coinciding features are removed from the target domains such that there is complete correspondence of the feature space between the source and the targets. In this way, not only the weights are transferred but also the model optimizer state. Only SICU, CCU-CTICU and CTICU qualify for this exercise as their input space is a subset of the target, Med-Surg ICU.
\begin{figure}[H]
    \centering
    \subfigure[Epochs]{
 \label{fig:epochs_fteicu}
        \includegraphics[width=.40\linewidth]{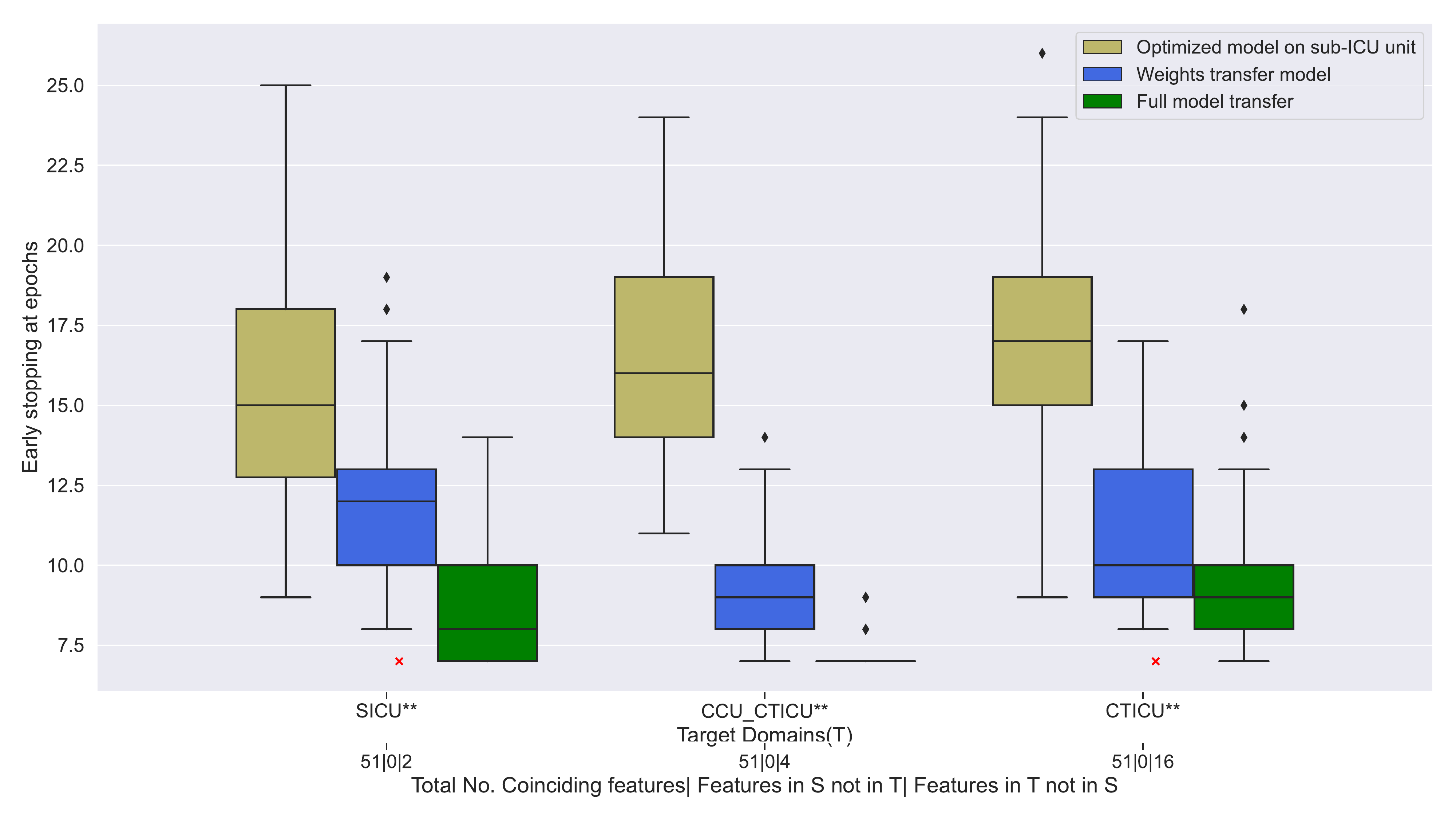}}
         \hspace{1.0cm}
    \subfigure[Error on test set]{
\label{fig:error_fteicu}
        \includegraphics[width=.40\linewidth]{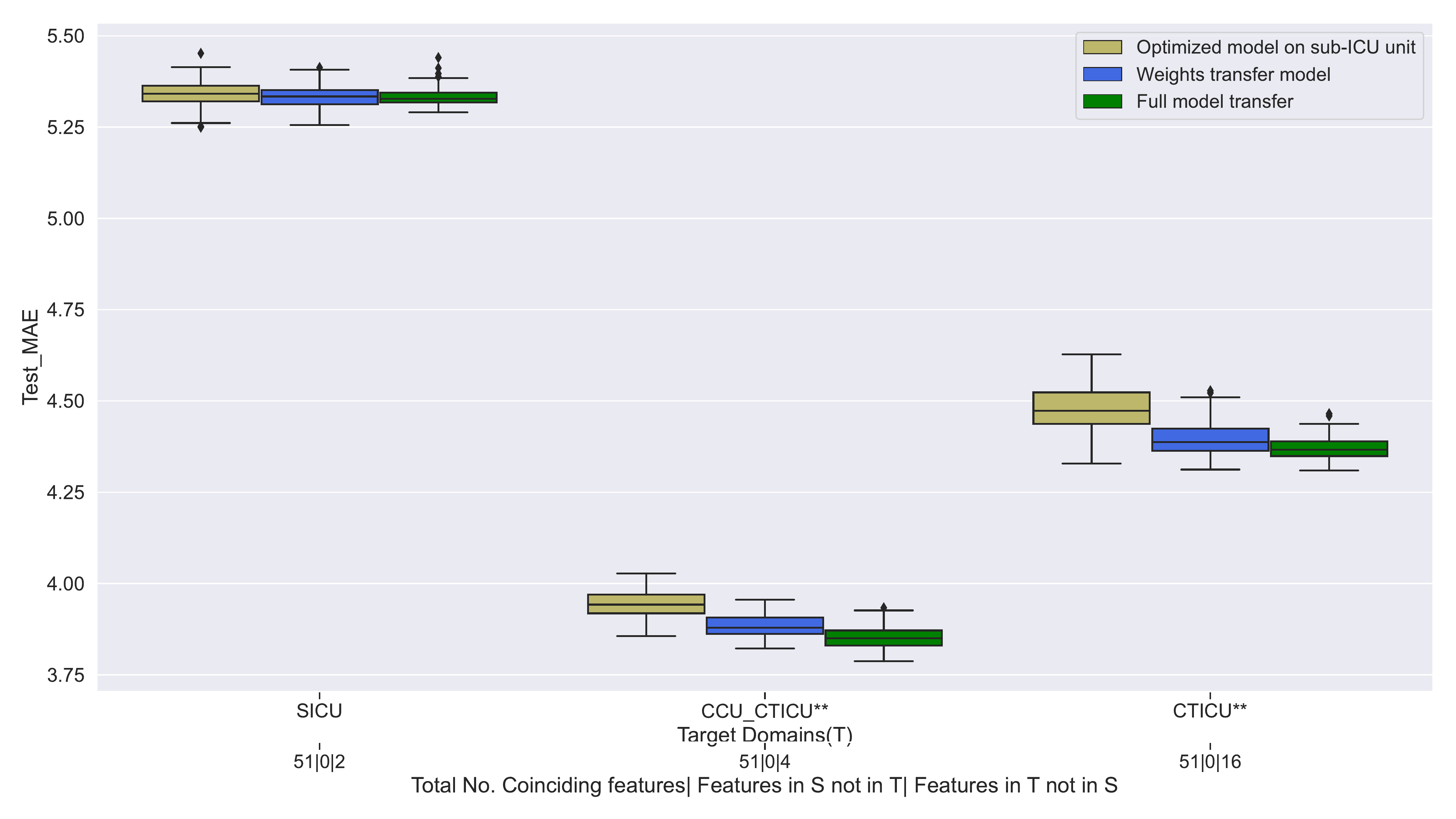}}
        \caption{Number of epochs and error measures when performing full model transfer. Tukey's HSD test is used for multiple comparisons\cite{abdi2010tukey}. Individual model (lemon green), weight transfer (blue) and full model transfer - including weights and optimizer state (dark green).}
    \label{fig:eicu_results_hps_full_model_transfer}
\end{figure}
Figure \ref{fig:eicu_results_hps_full_model_transfer} shows that if we had restricted the input space across all domains to be identical and continued training the source domain model on the target domains input data, thereby transferring not only the weights but also the optimizer state, model convergence would have occurred earlier than when transferring only the weights with an improvement in prediction accuracy. However, this would not be possible for all units because not all most recorded features in $S$ are found in all targets $T$.
\subsubsection{Weight Transfer with different learning rates}
\label{diff_lr}
Here, non-coinciding and coinciding features are trained simultaneously using different learning rates as explained in Section \ref{sec:diff_lr} and Eq (\ref{eqn:diff_lr}).
\begin{figure}[H]
    \centering
    \subfigure[Epochs]{
 \label{fig:epochs_difflreicu_csicu}        \includegraphics[width=.40\linewidth]{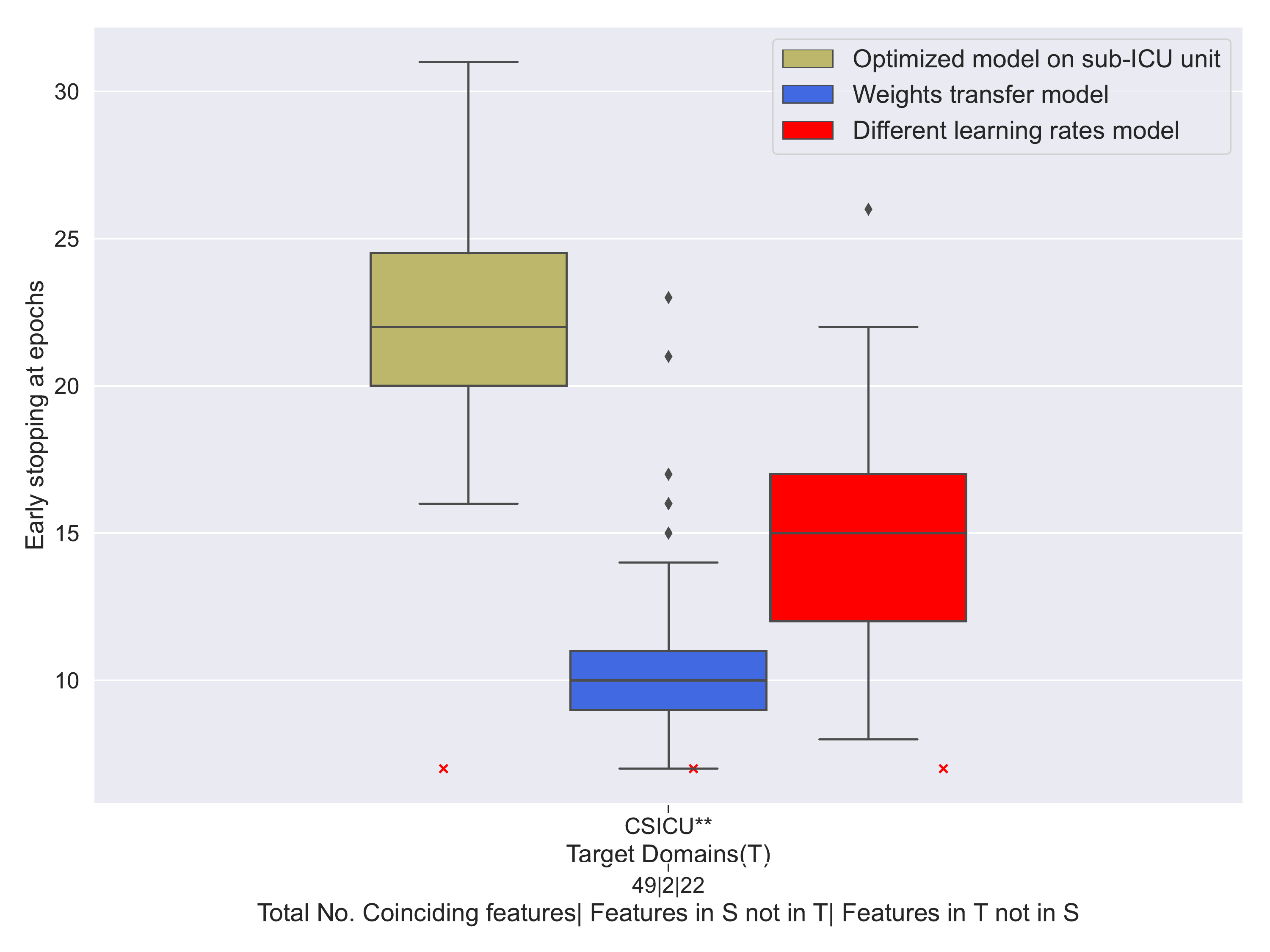}}
         \hspace{1.0cm}
    \subfigure[Error on test set]{
\label{fig:error_difflreicu_csicu}
        \includegraphics[width=.40\linewidth]{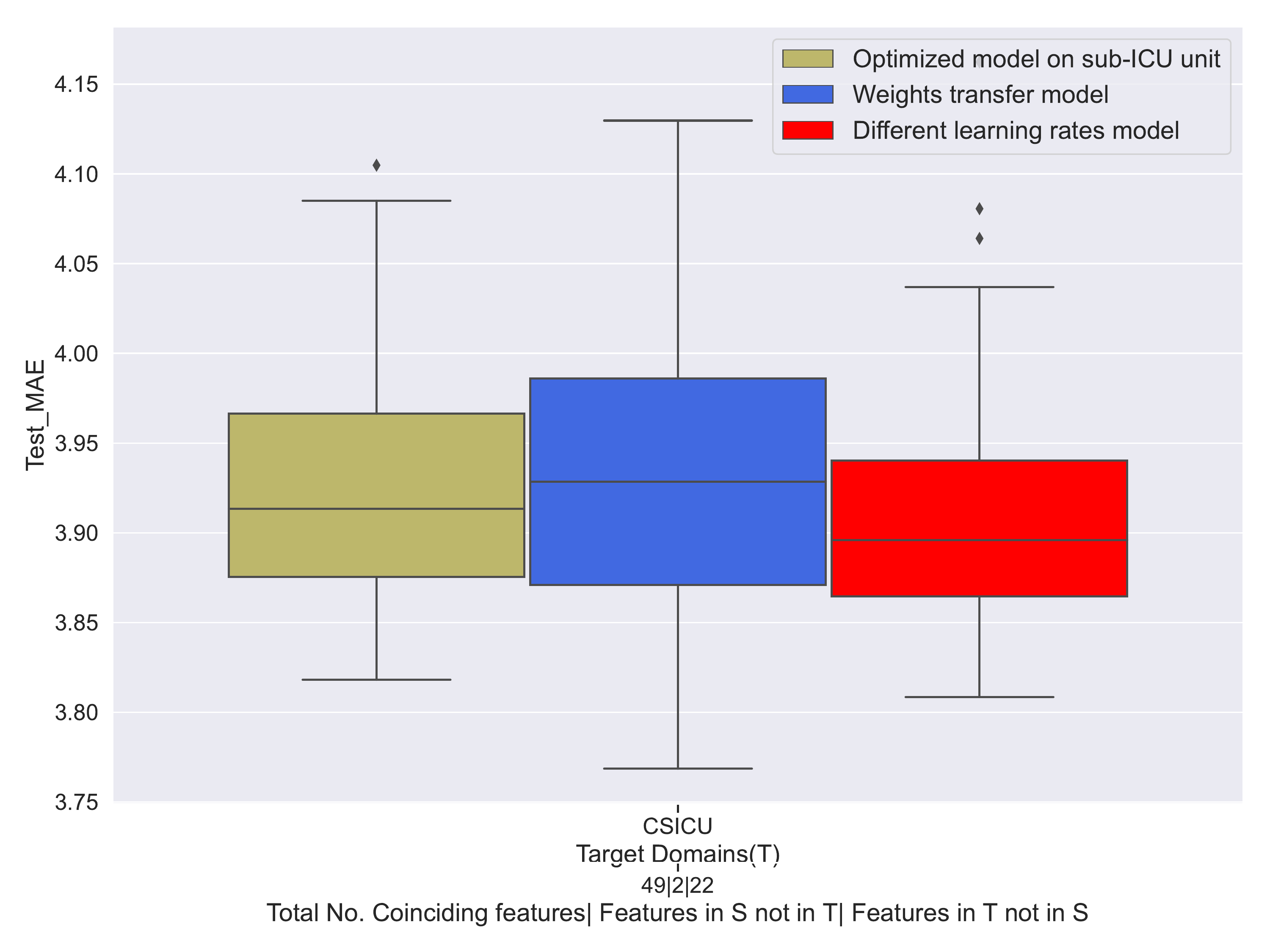}}
        \caption{Number of epochs and error measures on CSICU from eICU-CRD dataset after assigning different learning rates to different features for transfer learning. Tukey's HSD test is used for multiple comparisons \cite{abdi2010tukey}. Individual model (lemon green), weight transfer (blue) and different learning rates (red).}
    \label{fig:eicu_csicu_results_hps_diff_lr}
\end{figure}
Here it can be observed that indeed imposing discriminative learning produces a finer confidence interval (Figure \ref{fig:error_difflreicu_csicu}) for the prediction error while requiring more training time compared to weight transfer but still less than optimizing a full model (\ref{fig:epochs_difflreicu_csicu}).
\section{Discussion}
\label{sec: discussion}
The experiments in section \ref{sec:results} show that even partial weight transfer significantly improves prediction accuracy in a shorter time for most of the target domains. Furthermore, overall model importance is not affected. 
\subsection{Prediction error}
Section \ref{sec:results} showed that considering the ICU populations as a single homogeneous entity greatly reduces model utility by hospital management systems because each patient is assigned to a bed belonging to a specific unit. Thus the information of the LoS prediction at a more granular level of the unit is important. Moreover, tables \ref{tab:eICU_baseline_results} and \ref{tab:mimic_baseline_results} showed that an ``all stay" prediction model is not always an optimal estimation model for individuals groups. For e.g., CVICU patients in MIMIC-IV have a prediction error of 3.32 days, which is 1.79 days less than the former model. 
In eICU-CRD (Figures \ref{fig:eicu_epochs} and \ref{fig:eicu_error}), benefits are visible on populations such as CCU-CTICU (51,4), MICU (47,2) and CTICU (51,16) pairs of coinciding and non-coinciding features. However we notice that with (49,22), the prediction error increases for CSICU patients.
Though the difference in the mean prediction error for SICU population is not statistically significant, we observe that weight transfer model produces more confident results with a finer boxplot. \\
In MIMIC-IV, the main gain as observed in Figure \ref{fig:mimic_epochs_results} is that finer boxplots are obtained after weight transfer. For some targets, e.g., CVICU, a significant drop in prediction accuracy is observed in Figures \ref{fig:med_surg_source_mimic}, \ref{fig:micu_source}, \ref{fig:sicu_domain} and \ref{fig:cvicu_domain}. This could be associated to the fact that this unit has the highest number of input features. Thus, as target, a considerable number of non-coinciding features receive random weights (30 in Med-Surg ICU, 18 in MICU and 20 in SICU) that train alongside pre-trained weights.
\par The three further analyses carried out show that weight transfer has a significant positive impact on the model convergence and in several cases on the prediction error with at least, more confident error estimates.
\subsection{Computation time}
On average, the CPU time measured using python timer for the systematic Bayesian optimization procedure as explained in Appendix \ref{bayes_hps_tune} was 1 hour and 45 minutes (depending on the data size). Table \ref{tab:cpu_time_100_model_eicu} shows the execution times for some ICU units. $OP_m$: Optimized model on each ICU unit, $WT_m$: Weight transfer model, $FT_m$: Full model transfer.

\begin{table}[H]
\centering
\caption{CPU time in hours for re-training all models 100 times in eICU-CRD data.}
\begin{tabular}{cccc}
\toprule 
 ICU units & 100 $\times OP_m$&  100 $\times WT_m$ & 100 $\times FT_m$ \\ 
\midrule
 NICU & 2.24 & 1.40 & \\
MICU & 2.99 & 1.95 & \\
CSICU & 1.75 & 1.07 & \\
SICU & 2.87 & 2.39 & 1.86 \\
     \bottomrule
\end{tabular}
\label{tab:cpu_time_100_model_eicu}
\end{table}

Given our resources (see headline of Appendix \ref{hardware}) and the size of these units, a minimum of 1 hour was saved on units like MICU, CSICU taking into account the time needed for hyperparameter optimization. Optimization not done during TL since hyperparameters come from the source. Regarding MIMIC-IV data on Table \ref{tab:cpu_time_100_model_mimic}, over 2 hours are saved for target domain NS. As discussed before, weight transfer has a negative impact on CVICU as a target domain requiring more computation time.

\begin{table}[H]
\caption{CPU time in hours for re-training models 100 times in MIMIC-IV data}
\begin{center}
\begin{tabular}{c|cccc}
\multicolumn{1}{c}{$100 \times OP_m$ on $T_k$} & \multicolumn{4}{c} {$100 \times WT_m$ with $S_i$}\\
    \toprule
     NS & Med-Surg ICU & MICU & SICU & CVICU\\
     \multicolumn{5}{@{}c@{}}{\makebox[10cm]{\dashrule[black]}}\\
    3.63 & 1.66 & 1.78 & 1.85 & 1.59 \\
   \hline 
    Neuro-SICU & Med-Surg ICU & MICU & SICU & CVICU\\
    \multicolumn{5}{@{}c@{}}{\makebox[10cm]{\dashrule[black]}}\\
    3.44 & 2.32 & 2.26 & 2.21 & 2.60  \\
    \hline
    NI & Med-Surg ICU & MICU & SICU & CVICU\\  
    \multicolumn{5}{@{}c@{}}{\makebox[10cm]{\dashrule[black]}}\\
    2.75 & 2.11 & 2.23 & 2.28 & 2.29 \\
    \hline
    CVICU & Med-Surg ICU & MICU & SICU & CVICU\\  
    \multicolumn{5}{@{}c@{}}{\makebox[10cm]{\dashrule[black]}}\\
    1.75 & 1.99 & 2.18 & 2.10 & / \\
    \bottomrule
\end{tabular}
\end{center}
\label{tab:cpu_time_100_model_mimic}
\end{table}

\subsection{Insights into domain adaptation}
This section intends to discuss first, the reasons why domain adaptation works even when only partial information is transferred and secondly, the choice of the source domain.
\par
Regarding eICU-CRD data (Figure \ref{fig:eicu_results}), where the Med-Surg ICU unit is the only source domain, statistical significant improvements both in terms of early convergence and prediction error on all target populations except CSICU are observed. For CSICU, the error distribution appears to increase after partial weight transfer which could be due to the highest number of non-coinciding features (22) that receive random weights. As later shown in Figure \ref{fig:eicu_csicu_results_hps_diff_lr}, by assigning different learning rates, the error distribution is narrower (Figure \ref{fig:error_difflreicu_csicu}).
\par
In an attempt to investigate the effect of the choice of the source domain, MIMIC-IV data was used with four potential source domains. Figures \ref{fig:mimic_epochs_results} and \ref{fig:mimic_error_results} show that contrary to eICU-CRD where the source domain Med-Surg ICU occupies over 50\% of the data, this same domain returns the overall best performance when its weights are used to initiate training in the rest of the domains. Med-Surg ICU that contains both medical and surgical patients, that is a diverse patient population, has the greater impact on the other populations when used as source domain.  In terms of similarity between populations, Figure \ref{fig:sicu_domain} shows that the source domain SICU has the greatest impact on the target TSICU. However, instances when TL works or not could not all be explained. For e.g., in Figure \ref{fig:cvicu_domain}, CVICU has a slightly negative though non-significant impact on CCU. Though CVICU is a surgical unit type, and CCU a medical one, SICU which is also surgical appears to have a significant positive impact on CCU (Figure \ref{fig:sicu_domain}) with exactly the same 49 coinciding features. Thus pre-trained weights of CVICU used for total weight transfer do not improve learning as much as those of SICU.
\subsection{Effects of domain adaptation on model interpretability}
As shown in Figures \ref{fig:cticu_xai} and \ref{fig:tsicu_xai}, overall, the most important features which can be grouped under vital signs and lab parameters for early prediction of LoS are not affected by weight transfer though with slight changes in the order. Since the same groups of features appear as important before and after performing weight transfer.
\subsection{True hospital setting}
Though our work was not tested in a true hospital setting due to inability to access hospital data, we believe that transposing this to a true hospital setting where measures like the LILRANK \cite{lillrank2010demand} can be used to group or cluster departments based on their mode of functioning and other factors, our method can be applied at low cost within these clusters. Here the source can be chosen as a diverse population within the cluster and its weights used to initiate training to targets of the same cluster. This method can be of great benefit also for units with very small data sizes where their data will serve for fine-tuning pre-trained models rather than training a model from scratch. 
\section{Conclusions and Limitations}
\label{sec: conclusion}
In this work, domain adaptation is exploited to reuse knowledge learned from a source unto target domains by transferring learned weights from the trained source model.
By not restricting the input space such that it is identical across all units, we allow both shared and unit-specific information to be disseminated in the targets by fine-tuning both pre-trained (from the source) and random (unit-specific) weights. This resulted in statistically significant improvements in computation time as well as prediction accuracy for most of the targets.
However, we noticed that weight transfer was not always beneficial, especially when the target had a high number of non-coinciding features that receive random weights. By implementing discriminative learning and assigning different learning rates to the two feature groups (coinciding and non-coinciding), an improvement in the prediction error was noticed. In terms of feature importance, it appeared that the proposed approach maintains the overall importance by keeping the majority of important features though it displaces the order. Insights into this work showed that significant improvements in both prediction accuracy and computation time are observed when the source domain consists of a diverse population.
\par This work has a number of limitations. First, we couldn't fully understand all instances where weight transfer do not work. Secondly, when performing discriminative learning, a fixed factor of $\alpha=10^{-1}$ was used to reduce the learning rate for coinciding features. 
Thirdly, only time-varying features were used to predict LoS. Finally, the method was not evaluated in a true hospital setting. As future work, optimization of $\alpha$, the use of an adaptive learning rate and implementing mechanisms of data privacy, such as, differential privacy during weight transfer to prevent data leakage are envisaged.
\section*{CRediT authorship contribution statement}
\textbf{L.N.W.M}: Conceptualization, Methodology, Data curation, Model implementation, Writing - original draft \& editing. \textbf{N.M}: Methodology supervision. \textbf{E.N}: Methodology supervision \& draft review. \textbf{F.R}: Supervision, draft review \& editing. \textbf{B.DM}: Review. 
\section*{Declaration of competing Interest}
The authors declare that they have no known competing financial interests or personal relationships that could have appeared to influence the work reported in this paper.
\section*{Acknowledgments}
This work was supported by KU Leuven: Research Fund (projects C16/15/059, C3/19/053, C24/18/022, C3/20/117, C3I-21-00316), Industrial Research Fund (Fellowships 13-0260, IOFm/16/004, IOFm/20/002) and several Leuven Research and Development bilateral industrial projects;
Flemish Government Agencies:
FWO: EOS Project no G0F6718N (SeLMA), SBO project S005319N, Infrastructure project I013218N, TBM Project T001919N; PhD Grants (SB/1SA1319N, SB/1S93918, SB/1S1319N),
EWI: the Flanders AI Research Program
VLAIO: CSBO
(HBC.2021.0076) Baekeland PhD (HBC.20192204) and Innovation mandate (HBC.2019.2209)
European Commission: European Research Council under the European Union’s Horizon 2020 research and innovation programme (ERC Adv. Grant grant agreement No 885682);
Other funding: Foundation ‘Kom op tegen Kanker’, CM (Christelijke Mutualiteit) 
\appendix
\footnotesize
\section*{Appendix}
\section{Patient features extracted per ICU unit}
\label{sec:extracted_features}
\begin{enumerate}
    \item By using the pipeline by \cite{rocheteau2020temporal}, SQL queries were used to extract data tables (\textit{vitalperiodic}, \textit{vitalaperiodic}, \textit{respiratorycharting}, \textit{lab}, \textit{nursecharting} from eICU-CRD and \textit{labevents}, \textit{chartevents} from MIMIC-IV)  by imposing thresholds of presence of each feature in at least 25\% 13\% or 12.5\% of all adult patients. 
    \item Next, using the \textit{unit type} column from \textit{patient} table in eICU-CRD and \textit{first\_careunit} column from \textit{icustays} table in MIMIC-IV, stay ids were collected for each ICU unit type. 
    \item Extracted tables in 1. were then filtered on stays ids from 2. of each ICU unit for both datasets.  
    \item Again by modifying the pipeline by \cite{rocheteau2020temporal}, data curation and pre-processing was done per ICU unit type for both datasets one at a time. 
    \item In the pre-processing, all stays data tables for each unit, were merged, re-sampled hourly using the mean and scaled. From this, only features with at least 2 unique values for at least 30\% of the patients were retained for modelling. For each dataset and for each ICU unit type, final extracted features are grouped under the table they appear in the original database. 
    \item Steps 2., 3., 4. and 5. were also performed on the initial tables extracted in 1. without filtering per \textit{unit type} which constitutes the \textit{all stays} data. 
\end{enumerate}
\subsection{eICU data}
\subsection{MIMIC-IV data}
\begin{landscape}
\centering
\small
\begin{longtable}{|c|p{3cm}p{4cm}p{3cm}p{2cm}p{7cm}p{1cm}|}
\hline
Patients & \multicolumn{6}{|c|}{Source Tables}  \\
 \multirow{ 1}{*}{} & \textbf{\textit{vitalperiodic}} & \textbf{\textit{vitalaperiodic}} & \textbf{\textit{respiratorycharting}} & \textbf{\textit{lab}} & \textbf{\textit{nursecharting}} & \textbf{\textit{N/A}} 
  \\ \hline
\multirow{ 1}{*}{All Stays} & cvp, heartrate, st1, st2, st3, respiration, temperature, sao2  &noninvasivediastolic, noninvasivemean, noninvasivesystolic, systemicdiastolic, systemicmean, systemicsystolic &  &  FiO2, bedside glucose&  Glasgow coma score, Heart Rate, O2 L/\%, O2 Saturation, Pain Score/Goal,  Invasive BP, Non-Invasive BP, Respiratory Rate, Sedation Scale/Score/Goal, Temperature& Time in ICU
\\\hline
\multirow{ 1}{*}{Med Surg ICU} & cvp, heartrate, st1, st2, st3, respiration, temperature, sao2  &noninvasivediastolic, noninvasivemean, noninvasivesystolic, systemicdiastolic, systemicmean, systemicsystolic &  &  FiO2, bedside glucose&  Glasgow coma score, Heart Rate, O2 L/\%, O2 Saturation, Pain Score/Goal,  Invasive BP, Non-Invasive BP, Respiratory Rate, Temperature& Time in ICU
\\\hline
\multirow{ 1}{*}{SICU} & cvp, heartrate, st1, st2, st3, respiration, temperature, sao2  &noninvasivediastolic, noninvasivemean, noninvasivesystolic, systemicdiastolic, systemicmean, systemicsystolic &  &  FiO2, bedside glucose&  Glasgow coma score, Heart Rate, O2 L/\%, O2 Saturation, Pain Score/Goal,  Invasive BP, Non-Invasive BP, Respiratory Rate, Sedation Scale/Score/Goal, Temperature& Time in ICU
  \\ \hline
\multirow{ 1}{*}{NICU} & heartrate, st1, st2, st3, respiration, sao2  &noninvasivediastolic, noninvasivemean, noninvasivesystolic, systemicdiastolic, systemicmean, systemicsystolic & Total RR &  FiO2, bedside glucose&  Glasgow coma score, Heart Rate, O2 L/\%, O2 Saturation, Pain Score/Goal,  Invasive BP, Non-Invasive BP, Respiratory Rate, Sedation Scale/Score/Goal, Temperature& Time in ICU
\\\hline
\multirow{ 1}{*}{MICU} & heartrate, st1, st2, st3, respiration, sao2  &noninvasivediastolic, noninvasivemean, noninvasivesystolic, systemicdiastolic, systemicmean, systemicsystolic & &  FiO2, bedside glucose&  Glasgow coma score, Heart Rate, O2 L/\%, O2 Saturation, Pain Score/Goal,  Invasive BP, Non-Invasive BP, Respiratory Rate, Sedation Scale/Score/Goal, Temperature& Time in ICU
\\\hline
\multirow{ 1}{*}{CTICU} & cvp, heartrate, st1, st2, st3, respiration, temperature, sao2  &noninvasivediastolic, noninvasivemean, noninvasivesystolic, systemicdiastolic, systemicmean, systemicsystolic &  &  FiO2, bedside glucose, HCO3, Hct, Hgb, pH, paCO2, paO2, potassium &  Glasgow coma score, Heart Rate, O2 L/\%, O2 Saturation, Pain Score/Goal, Invasive BP, Non-Invasive BP, Respiratory Rate, Temperature, Bedside Glucose & Time in ICU
\\\hline
\multirow{ 1}{*}{CICU} & cvp, heartrate, st1, st2, st3, respiration, sao2  & noninvasivediastolic, noninvasivemean, noninvasivesystolic, systemicdiastolic, systemicmean, systemicsystolic &PEEP & FiO2, bedside glucose &  Glasgow coma score, Heart Rate, O2 L/\%, O2 Saturation, Pain Score/Goal, Non-Invasive BP, Respiratory Rate, Sedation Scale/Score/Goal, Temperature & Time in ICU
\\\hline
\multirow{ 1}{*}{CCU-CTICU} & cvp, heartrate, st1, st2, st3, respiration, temperature, sao2  & noninvasivediastolic, noninvasivemean, noninvasivesystolic, systemicdiastolic, systemicmean, systemicsystolic & & FiO2, bedside glucose, potassium &  Glasgow coma score, Heart Rate, O2 L/\%, O2 Saturation, Pain Score/Goal, Invasive BP, Non-Invasive BP, Respiratory Rate, Sedation Scale/Score/Goal, Temperature & Time in ICU
\\\hline
\multirow{ 1}{*}{CSICU} & cvp, heartrate, st1, st2, st3, respiration, sao2  &noninvasivediastolic, noninvasivemean, noninvasivesystolic, systemicdiastolic, systemicmean, systemicsystolic & Total RR, PEEP, Peak Insp. Pressure, TV/kg IBW, Mean Airway Pressure, Tidal Volume (set), Vent Rate, Exhaled MV, LPM O2 &  FiO2, bedside glucose, potassium &  Glasgow coma score, Heart Rate, O2 L/\%, O2 Saturation, Pain Score/Goal, Invasive BP, Non-Invasive BP, Respiratory Rate, Temperature, Bedside Glucose & Time in ICU
\\\hline
\caption{List of Features extracted per ICU unit in eICU-CRD.}
\end{longtable}
\label{eICU_feature_Set}
\end{landscape}

\begin{landscape}
\centering
\small
\begin{longtable}{|c|p{6cm}p{12cm}r|}
\hline
Patients & \multicolumn{3}{|c|}{Source Tables}  \\
 \multirow{1}{*}{} & \textbf{\textit{labevents}} & \textbf{\textit{chartevents}} &  \textbf{\textit{N/A}} 
  \\ \hline
\multirow{1}{*}{All Stays} & Base Excess, Calculated Total CO2, Glucose, Hematocrit, Hemoglobin, pCO2, pH, pO2 & Activity / Mobility (JH-HLM), Arterial Blood Pressure diastolic, Arterial Blood Pressure mean, Arterial Blood Pressure systolic, GCS - Eye Opening, GCS - Motor Response, GCS - Verbal Response, Glucose finger stick (range 70-100), Heart Rate, Inspired O2 Fraction, Mean Airway Pressure, Minute Volume, Non Invasive Blood Pressure diastolic, Non Invasive Blood Pressure mean,
Non Invasive Blood Pressure systolic, O2 saturation pulseoxymetry, PEEP set, Respiratory Rate, Richmond-RAS Scale, Strength L Arm, Strength L Leg, Strength R Arm, Strength R Leg, Temperature Fahrenheit, Tidal Volume (observed) & hour
 \\ \hline
\multirow{1}{*}{MICU} &  Glucose, Hematocrit, Hemoglobin, pH & Activity / Mobility (JH-HLM), Arterial Blood Pressure diastolic, Arterial Blood Pressure mean, Arterial Blood Pressure systolic, GCS - Eye Opening, GCS - Motor Response, GCS - Verbal Response, Glucose finger stick (range 70-100), Heart Rate, Inspired O2 Fraction, Mean Airway Pressure, Minute Volume, Non Invasive Blood Pressure diastolic, Non Invasive Blood Pressure mean,
Non Invasive Blood Pressure systolic, O2 saturation pulseoxymetry, PEEP set, Respiratory Rate, Richmond-RAS Scale, Strength L Arm, Strength L Leg, Strength R Arm, Strength R Leg, Temperature Fahrenheit, Tidal Volume (observed) & hour
 \\ \hline
\multirow{1}{*}{CVICU} & Base Excess, Calculated Total CO2, Free Calcium, Glucose, Hematocrit, Hematocrit, Calculated, Hemoglobin, Lactate, Potassium, Whole Blood, pCO2, pH, pO2 & Activity / Mobility (JH-HLM), Arterial Blood Pressure diastolic, Arterial Blood Pressure mean, Arterial Blood Pressure systolic, GCS - Eye Opening, GCS - Motor Response, GCS - Verbal Response, Glucose finger stick (range 70-100), Heart Rate, Inspired O2 Fraction, Mean Airway Pressure, Minute Volume, Non Invasive Blood Pressure diastolic, Non Invasive Blood Pressure mean,
Non Invasive Blood Pressure systolic, O2 saturation pulseoxymetry, PEEP set, Respiratory Rate, Richmond-RAS Scale, Strength L Arm, Strength L Leg, Strength R Arm, Strength R Leg, Temperature Fahrenheit, Tidal Volume (observed), Ventilator Mode & hour
 \\ \hline
\multirow{1}{*}{Med-Surg ICU} & Glucose, Hematocrit, & Activity / Mobility (JH-HLM), Arterial Blood Pressure diastolic, Arterial Blood Pressure mean, Arterial Blood Pressure systolic, GCS - Eye Opening, GCS - Motor Response, GCS - Verbal Response, Glucose finger stick (range 70-100), Heart Rate, Inspired O2 Fraction, Non Invasive Blood Pressure diastolic, Non Invasive Blood Pressure mean,
Non Invasive Blood Pressure systolic, O2 saturation pulseoxymetry, Respiratory Rate, Richmond-RAS Scale, Strength L Arm, Strength L Leg, Strength R Arm, Strength R Leg, Temperature Fahrenheit & hour
 \\ \hline
\multirow{1}{*}{SICU} & Glucose, Hematocrit, Hemoglobin & Activity / Mobility (JH-HLM), Arterial Blood Pressure diastolic, Arterial Blood Pressure mean, Arterial Blood Pressure systolic, GCS - Eye Opening, GCS - Motor Response, GCS - Verbal Response, Glucose finger stick (range 70-100), Heart Rate, Inspired O2 Fraction, Mean Airway Pressure, Minute Volume, Non Invasive Blood Pressure diastolic, Non Invasive Blood Pressure mean,
Non Invasive Blood Pressure systolic, O2 saturation pulseoxymetry, PEEP set, Respiratory Rate, Richmond-RAS Scale, Strength L Arm, Strength L Leg, Strength R Arm, Strength R Leg, Temperature Fahrenheit, Tidal Volume (observed) & hour
 \\ \hline
\multirow{1}{*}{TSICU} & Base Excess, Calculated Total CO2, Glucose, Hematocrit, Hemoglobin, pCO2, pH, pO2 & Activity / Mobility (JH-HLM), Arterial Blood Pressure diastolic, Arterial Blood Pressure mean, Arterial Blood Pressure systolic, GCS - Eye Opening, GCS - Motor Response, GCS - Verbal Response, Glucose finger stick (range 70-100), Heart Rate, Inspired O2 Fraction, Mean Airway Pressure, Minute Volume, Non Invasive Blood Pressure diastolic, Non Invasive Blood Pressure mean,
Non Invasive Blood Pressure systolic, O2 saturation pulseoxymetry, PEEP set, Respiratory Rate, Richmond-RAS Scale, Strength L Arm, Strength L Leg, Strength R Arm, Strength R Leg, Temperature Fahrenheit, Tidal Volume (observed), Ventilator Mode & hour
 \\ \hline
\multirow{1}{*}{CCU} & Glucose, Hematocrit, Hemoglobin & 
Activity / Mobility (JH-HLM), Arterial Blood Pressure diastolic, Arterial Blood Pressure mean, Arterial Blood Pressure systolic, GCS - Eye Opening, GCS - Motor Response, GCS - Verbal Response, Glucose finger stick (range 70-100), Heart Rate, Inspired O2 Fraction, Non Invasive Blood Pressure diastolic, Non Invasive Blood Pressure mean,
Non Invasive Blood Pressure systolic, O2 saturation pulseoxymetry, Respiratory Rate, Richmond-RAS Scale, Strength L Arm, Strength L Leg, Strength R Arm, Strength R Leg, Temperature Fahrenheit & hour
 \\ \hline
\multirow{1}{*}{Neuro-SICU} & & H, I, L, Activity / Mobility (JH-HLM), Arterial Blood Pressure diastolic, Arterial Blood Pressure mean, Arterial Blood Pressure systolic, GCS - Eye Opening, GCS - Motor Response, GCS - Verbal Response, Glucose finger stick (range 70-100), Heart Rate, Inspired O2 Fraction, Mean Airway Pressure, Minute Volume, Non Invasive Blood Pressure diastolic, Non Invasive Blood Pressure mean,
Non Invasive Blood Pressure systolic, O2 saturation pulseoxymetry, PEEP set, Respiratory Rate, Richmond-RAS Scale, Strength L Arm, Strength L Leg, Strength R Arm, Strength R Leg, Temperature Fahrenheit, Tidal Volume (observed) & hour
 \\ \hline
\multirow{1}{*}{NI} & & Activity / Mobility (JH-HLM), Arterial Blood Pressure diastolic, Arterial Blood Pressure mean, Arterial Blood Pressure systolic, GCS - Eye Opening, GCS - Motor Response, GCS - Verbal Response, Glucose finger stick (range 70-100), Heart Rate, Non Invasive Blood Pressure diastolic, Non Invasive Blood Pressure mean,
Non Invasive Blood Pressure systolic, O2 saturation pulseoxymetry, Respiratory Rate, Richmond-RAS Scale, Strength L Arm, Strength L Leg, Strength R Arm, Strength R Leg, Temperature Fahrenheit & hour
 \\ \hline
\multirow{1}{*}{NS} & & Activity / Mobility (JH-HLM), Arterial Blood Pressure diastolic, Arterial Blood Pressure mean, Arterial Blood Pressure systolic, GCS - Eye Opening, GCS - Motor Response, GCS - Verbal Response, Glucose finger stick (range 70-100), Heart Rate, Non Invasive Blood Pressure diastolic, Non Invasive Blood Pressure mean,
Non Invasive Blood Pressure systolic, O2 saturation pulseoxymetry, Respiratory Rate, Richmond-RAS Scale, Strength L Arm, Strength L Leg, Strength R Arm, Strength R Leg, Temperature Fahrenheit & hour
\\ \hline
\caption{List of Features extracted per ICU unit in MIMIC-IV.}
\end{longtable}
\end{landscape}
\section{Implementation details}
\label{hardware}
All algorithms were written in Python language. All experiments were performed on a 128GB RAM computer with Intel(R) Core(TM) i9-10900X CPU processor and NVIDIA GeForce RTX 2080 Ti GPU.

\subsection{Hyperparameter Search Methodology and Implementation Details}
\label{bayes_hps_tune}
The hyperparameter search space for all ICU units is given in the table below. An asterisk * is indicated for those in which the space was enlarged for particular units.
\begin{table}[h]
    \caption{Hyperparameters search space and corresponding scale. No. Hidden units was enlarged to the interval [4 - 512] for Neuro-Stepdown ICU patients in MIMIC-IV and the learning rate to [1$e^{-5}-1e^{-2}$] for Medical-Surgical ICU patients in MIMIC-IV}
    \centering
    \begin{tabular}{lcl}
    \toprule
       Hyperparameter  & Search Space Interval & Scale  \\
       \midrule
         No.Hidden Layers &[1 - 2]& Linear\\
         No.Hidden units* & [8 - 512] &$\log_2$\\
         Learning rate* &[1$e^{-4}-1e^{-2}$]& $\log_{10}$\\
         Dropout &[0.1 - 0.5]& Linear \\
         Batch Size &[4 - 512]&$\log_2$\\
         \bottomrule
    \end{tabular}
    \label{tab:hps}
\end{table}\\
Hyperparameter search was performed in a systematic manner for each of the patient populations using the Bayesian Optimizer from KerasTuner \cite{o2019keras, omalley2019kerastuner}. Essentially, 
Bayesian Optimization makes use of both a prior function and an acquisition function. The former is used as a surrogate to obtain estimates of the objective function and the latter, that measures the evaluation of the objective function at a new point and proposes next candidate points within the search space \cite{snoek2012practical}.
When using the Bayesian optimization, we minimize the validation loss, that is the, mean squared logarithmic error on the validation dataset. As explained before, the candidate hyperparameters values used at each trial are proposed depending on the performance of the previously chosen values. We noticed that orienting the search direction seriously affects the number of hidden units. Therefore, the hidden units space was split into two resulting in a three-step procedure as follows;
\begin{enumerate}
    \item Firstly, the search space involving all hyperparameters except the batch size and No. hidden units in the interval [8 - 64] was used. 
    \item Secondly, the previous step was repeated with No. hidden units in the interval [64 - 512].
    \item In the third step, the refined space following model performance was used and hyperparameter search was repeated. 
    \item The best hyperparameters from step three were then used to fit the model using early stopping to prevent overfitting. By monitoring the validation loss over six epochs, early stopping occurs if this doesn't further decrease by at least 0.5\%. 
\end{enumerate}
In steps 1 and 2, ten trials were done with two executions per trial. In step 3, ten trials were done with three executions per trial. By performing multiple executions per trial ensures that the reported hyperparameter values return the lowest and most stable average error over all trials.
\section{Loss Curves}
Additional loss curves from Section \ref{hyperpar_optim}
\begin{figure}[H]
    \centering
    \subfigure[CSICU patients]{
\label{fig:csicu loss}
        \includegraphics[width=.40\linewidth]{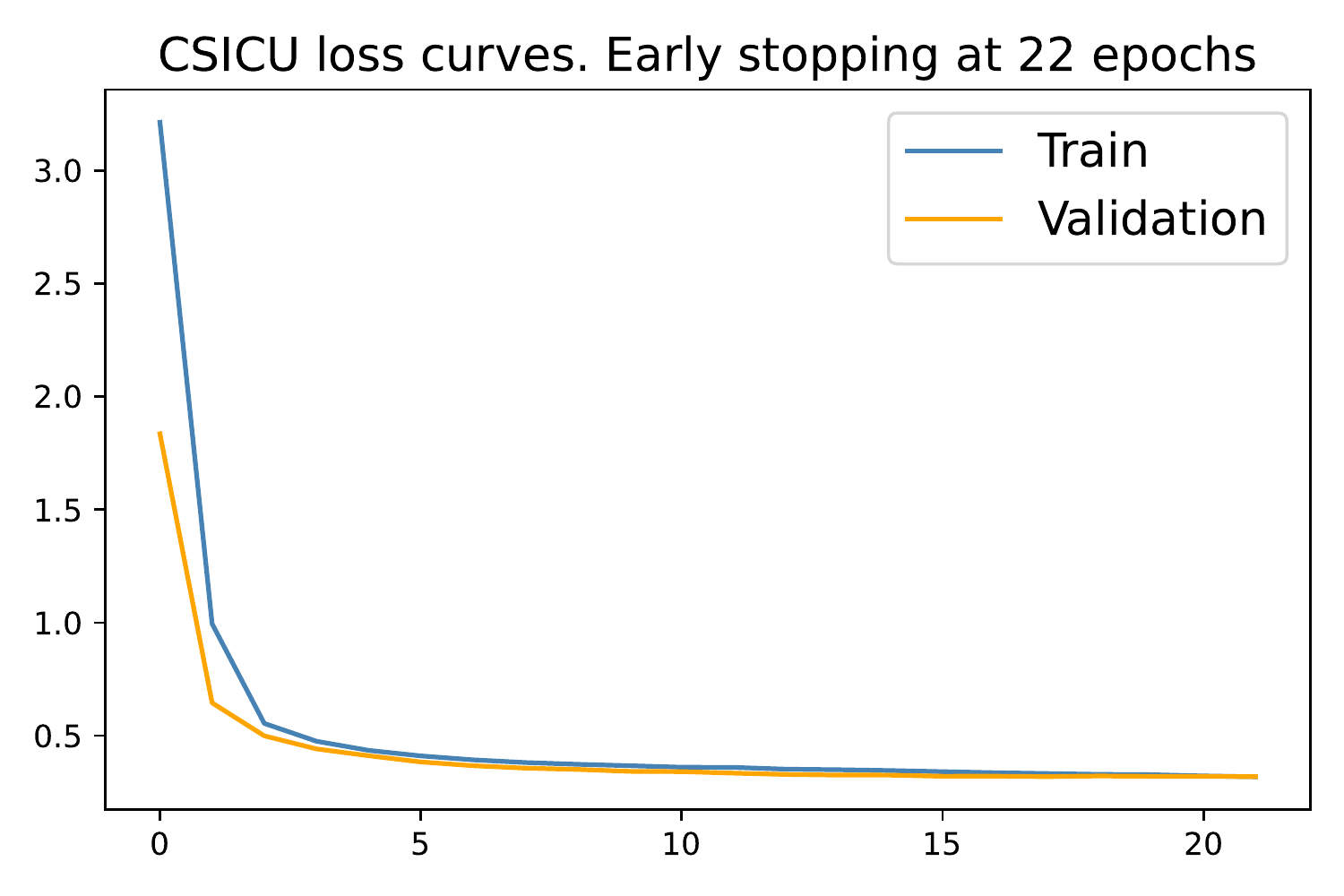}}
 \hspace{1.0cm}        
\subfigure[SICU patients]{
\label{fig:sicu_loss}
\includegraphics[width=.40\linewidth]{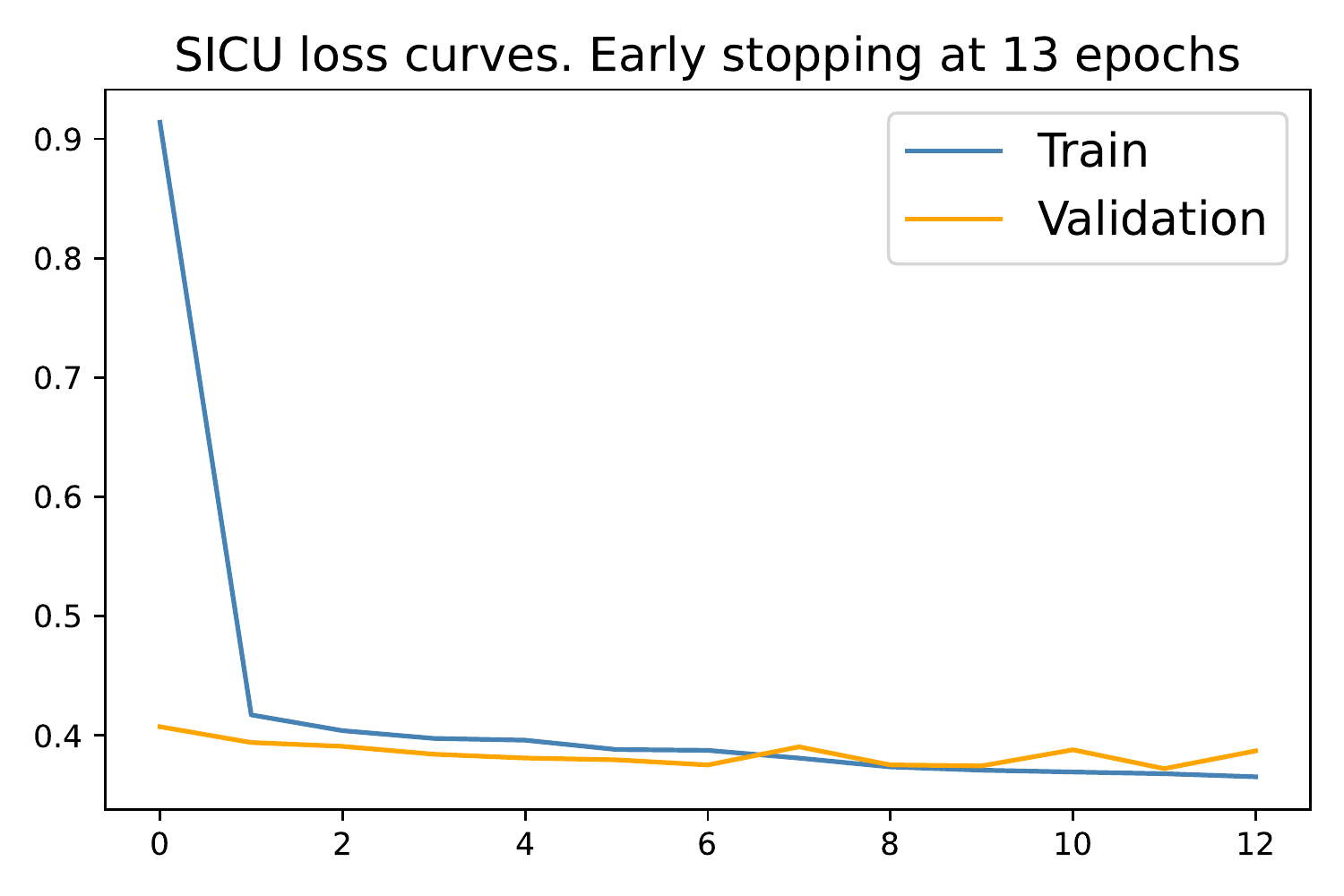}}
\caption{Loss curves obtained after training CSICU and SICU in eICU-CRD data with optimized hyperparameters}
\label{fig:appendix_los_eicu}
\end{figure}

\begin{figure}[H]
    \centering
    \subfigure[MICU patients]{
\label{fig:micu loss}
        \includegraphics[width=.40\linewidth]{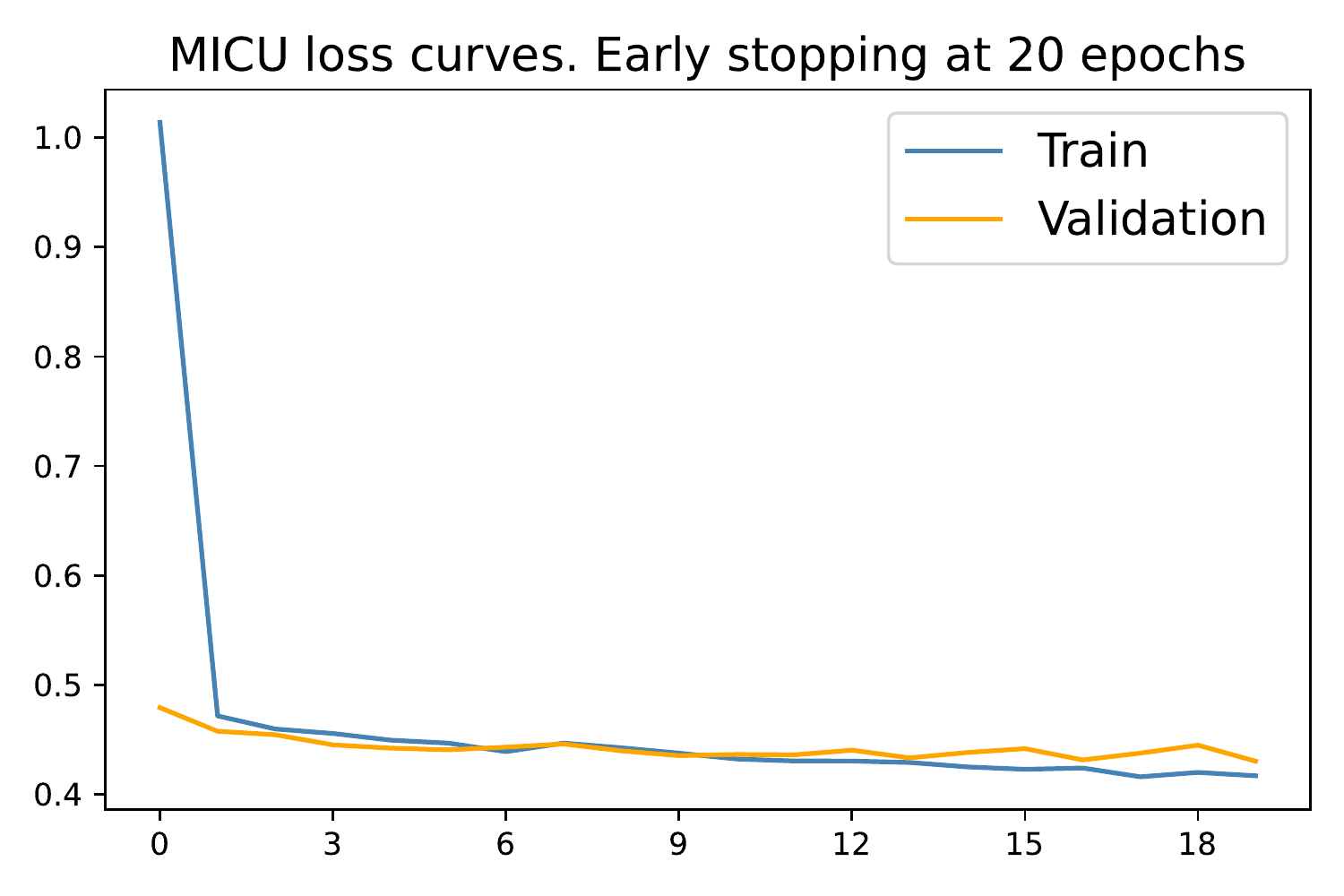}}
 \hspace{1.0cm}        
\subfigure[SICU patients]{
\label{fig:sicu_loss_mimic}
\includegraphics[width=.40\linewidth]{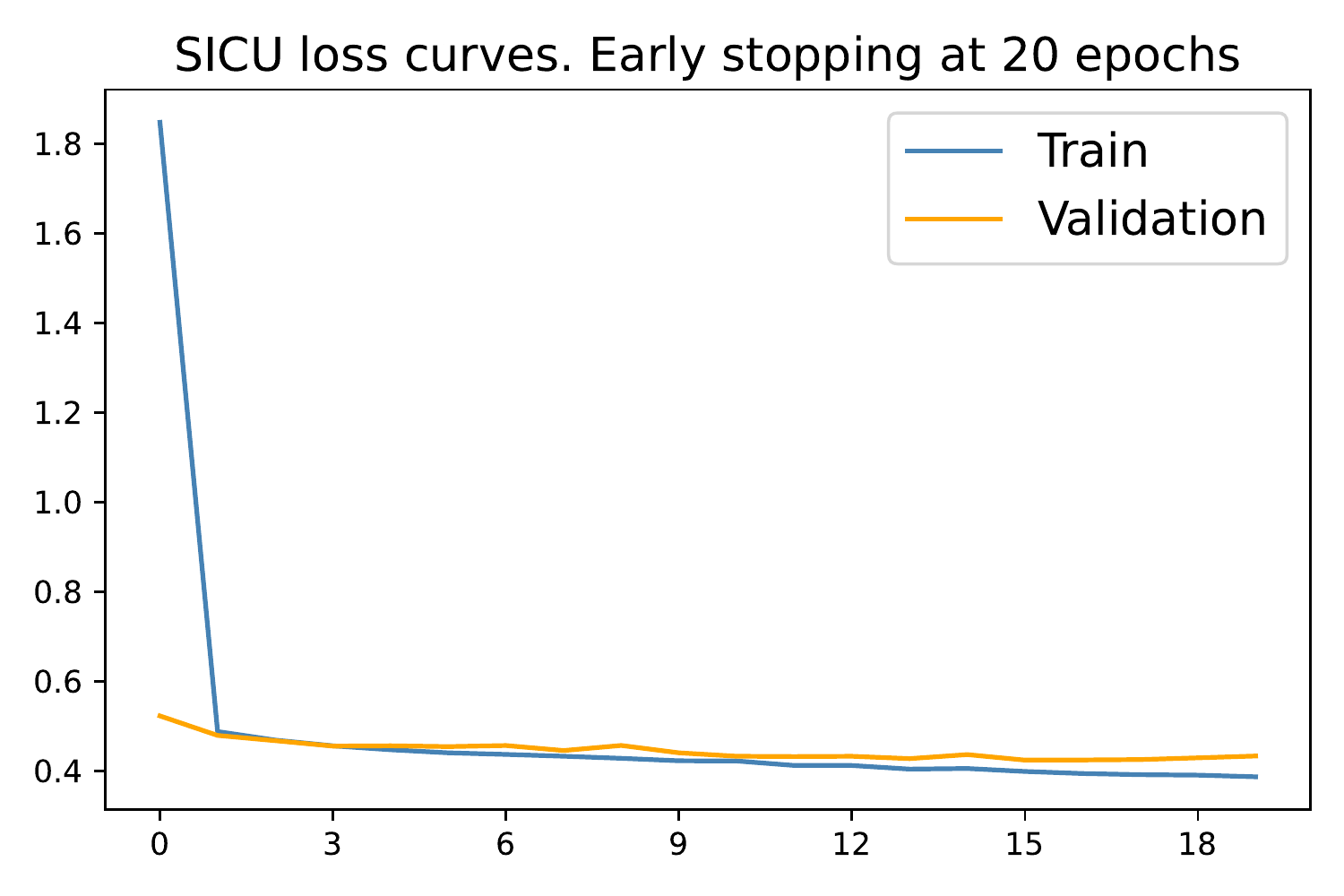}}
\caption{Loss curves obtained after training MICU and SICU in MIMIC-IV data with optimized hyperparameters as in Section \ref{hyperpar_optim}.}
\label{fig:appendix_los_mimic}
\end{figure}

\section{Multiple comparisons of Means using Tukey test}
Complementary results for Figure \ref{fig:eicu_results_hps_full_model_transfer} and \ref{fig:eicu_csicu_results_hps_diff_lr}.
\begin{table}[H]
\caption{SICU: Multiple comparison of Means for epochs and Test MAE using Tukey HSD \cite{abdi2010tukey}, $\alpha = 5\%$. OM: Optimized Model on each sub-ICU unit, WTM: Weight Transfer Model, FTM: Full Transfer Model}
\begin{center}
\begin{tabular}{cccccccc}
    \toprule
    & group 1 & group 2 & mean diff & p-adj & lower & upper & reject H$_0$\\
    \midrule
   \multirow{2}{*}{Epochs} & OM & WTM & -3.6331 & 0.001 & -4.5746 & -2.6917 & True\\
    & FTM & OM & 6.3301 & 0.001 & 5.3886 & 7.2716 & True\\
    & FTM & WTM & -2.697 & 0.001 & 1.7531 & 3.6408 & True\\
    \hline
   \multirow{2}{*}{Test MAE} & OM & WTM &-0.0071 & 0.2555 & -0.0177 & 0.0035 & False\\
    & FTM & OM & 0.0095& 0.0891 & -0.0011 & 0.02 & False\\
    & FTM & WTM & 0.0024 & 0.8418 & -0.0082 & 0.013 & False\\
    \bottomrule
\end{tabular}
\end{center}
\label{tab:tukey eicu ft sicu}
\end{table}

\begin{table}[H]
\caption{CCU-CTICU: Multiple comparison of Means for epochs and Test MAE using Tukey HSD, $\alpha = 5\%$. OM: Optimized Model on each sub-ICU unit, WTM: Weight Transfer Model, FTM: Full Transfer Model}
\begin{center}
\begin{tabular}{cccccccc}
    \toprule
    & group 1 & group 2 & mean diff & p-adj & lower & upper & reject H$_0$\\
    \midrule
    \multirow{2}{*}{Epochs} & OM & WTM &-7.07 & 0.001 & -7.7585 & -6.3815 & True\\
    & FTM & OM & 9.04 & 0.001 & 8.3515 & 9.7285 & True\\
    & FTM & WTM & 1.97 & 0.001 & 1.2815 & 2.6585 & True\\
    \hline
   \multirow{2}{*}{Test MAE} & OM & WTM &-0.0591 & 0.001 & -0.0702 & -0.048 & True\\
    & FTM & OM & 0.0922 & 0.001 & 0.0812 & 0.1033 & True\\
    & FTM & WTM & 0.0331 & 0.001 & 0.022 & 0.0442 & True\\
    \bottomrule
\end{tabular}
\end{center}
\label{tab:tukey eicu ft ccu_cticu}
\end{table}
\begin{table}[H]
\caption{CTICU: Multiple comparison of Means for epochs and Test MAE using Tukey HSD, $\alpha = 5\%$. OM: Optimized Model on each sub-ICU unit, WTM: Weight Transfer Model, FTM: Full Transfer Model}
\begin{center}
\begin{tabular}{cccccccc}
    \toprule
    & group 1 & group 2 & mean diff & p-adj & lower & upper & reject H$_0$\\
    \midrule
   \multirow{2}{*}{Epochs} & OM & WTM &-6.2298 & 0.001 & -7.0824 & -5.3772 & True\\
    & FTM & OM & 7.8012 & 0.001 & 6.9487 & 8.6538 & True\\
    & FTM & WTM & 1.5714 & 0.001 & 0.7146 & 2.42283 & True\\
    \hline
   \multirow{2}{*}{Test MAE} & OM & WTM &-0.0829 & 0.001 & -0.0992 & -0.0666 & True\\
    & FTM & OM & 0.1079 & 0.001 & 0.00916 & 0.1242 & True\\
    & FTM & WTM & 0.0249 & 0.0011 & 0.0086 & 0.0413 & True\\
    \bottomrule
\end{tabular}
\end{center}
\label{tab:tukey eicu ft cticu}
\end{table}

\begin{table}[H]
\caption{CSICU: Multiple comparison of Means for epochs and Test MAE using Tukey HSD, $\alpha = 5\%$. OM: Optimized Model on each sub-ICU unit, WTM: Weight Transfer Model, Diff$\_$LR: Different Learning Rates Model}
\begin{center}
\begin{tabular}{cccccccc}
    \toprule
    & group 1 & group 2 & mean diff & p-adj & lower & upper & reject H$_0$\\
    \midrule
   \multirow{2}{*}{Epochs} & Diff$\_$LR & OM & 7.7071 & 0.001 & 6.6032 & 8.811 & True\\
    & Diff$\_$LR & WTM & -4.0707 & 0.001 & -5.1746 & -2.9668 & True\\
    & OM & WTM & -11.7778 & 0.001 & -12.8817 & -10.6739 & True\\
    \hline
   \multirow{2}{*}{Test MAE} & Diff$\_$LR & OM & 0.0173 & 0.2146 & -0.0069 & 0.0415 & False\\
    & Diff$\_$LR & WTM & 0.0214  & 0.0961 & -0.0029 & 0.0456 & True\\
    & OM & WTM & 0.0041 & 0.9 & -0.0201 & 0.0283 & False\\
    \bottomrule
\end{tabular}
\end{center}
\label{tab:tukey eicu_csicu}
\end{table}

\printbibliography
\end{document}